\documentclass{article}
\usepackage[preprint]{neurips_2021}
\usepackage[utf8]{inputenc} 
\usepackage[T1]{fontenc}    
\usepackage{hyperref}       
\usepackage{url}            
\usepackage{booktabs}       
\usepackage{amsfonts}       
\usepackage{nicefrac}       
\usepackage{microtype}      
\usepackage{xcolor}         
\usepackage{graphicx}
\usepackage{amsmath}
\usepackage{amsthm}
\usepackage{algorithm}
\usepackage{algorithmic}
\usepackage{subfigure}
\usepackage{enumerate}
\usepackage{array}
\usepackage{wrapfig}
\title{State-based Episodic Memory \\for Multi-Agent Reinforcement Learning}

\author{%
  Xiao Ma \\
Department of Computer Science and Technology\\
Nanjing University\\
  \texttt{max@lamda.nju.edu.cn} \\
  \And
  Wu-Jun Li\\
  National Key Laboratory for Novel Software Technology\\
Department of Computer Science and Technology\\
Nanjing University\\
  \texttt{liwujun@nju.edu.cn}\\
}

\begin{document}

\maketitle

\begin{abstract}
Multi-agent reinforcement learning~(MARL) algorithms have made promising progress in recent years by leveraging the centralized training and decentralized execution~(CTDE) paradigm. However, existing MARL algorithms still suffer from the sample inefficiency problem. In this paper, we propose a simple yet effective approach, called \underline{s}tate-based \underline{e}pisodic \underline{m}emory~(SEM), to improve sample efficiency in MARL. SEM adopts episodic memory~(EM) to supervise the centralized training procedure of CTDE in MARL. To the best of our knowledge, SEM is the first work to introduce EM into MARL. We can theoretically prove that, when using for MARL, SEM has lower space complexity and time complexity than \underline{s}tate and \underline{a}ction based \underline{EM}~(SAEM), which is originally proposed for single-agent reinforcement learning. Experimental results on StarCraft multi-agent challenge~(SMAC) show that introducing episodic memory into MARL can improve sample efficiency and SEM can reduce storage cost and time cost compared with SAEM.

\end{abstract}

\section{Introduction}
\label{Introduction}
Reinforcement learning~(RL) has achieved promising success in a variety of challenging domains, including game playing~\cite{DBLP:journals/nature/MnihKSRVBGRFOPB15} and robotics~\cite{DBLP:journals/corr/LillicrapHPHETS15}. There are multiple agents to collaboratively make sequential decisions in many real-world applications, such as autonomous driving~\cite{DBLP:journals/corr/Shalev-ShwartzS16a}, intelligent robotic control~\cite{DBLP:journals/ijrr/KoberBP13}, and game playing~\cite{DBLP:journals/corr/abs-1912-06680,DBLP:conf/atal/SamvelyanRWFNRH19}. Therefore, multi-agent reinforcement learning~(MARL) has attracted much attention. In MARL, each agent has its partial observations and chooses its actions in a shared environment with other agents' interactions. This complex interaction model causes many challenges, such as instability, sample inefficiency, the moving target problem (non-stationarity), the exponential growth of the agents' action space. 

Early MARL methods, such as independent Q-learning~\cite{tan1993multi}, adopt the decentralized policy for training. In these methods, each agent takes action independently and treats other agents as part of the environment. Since other agents' policies will change, leading to a non-stationary environment, these decentralized policy based methods suffer from sample inefficiency and instability problems. To solve these problems caused by decentralized policies, researchers recently propose to use the paradigm called centralized training and decentralized execution~(CTDE)~\cite{DBLP:journals/jair/OliehoekSV08}. Benefiting from CTDE, value-decomposed MARL algorithms~\cite{DBLP:conf/nips/LoweWTHAM17,DBLP:conf/atal/SunehagLGCZJLSL18,DBLP:conf/icml/RashidSWFFW18,DBLP:conf/aaai/FoersterFANW18,DBLP:conf/nips/RashidFPW20,DBLP:journals/corr/abs-2008-01062} have been proposed in recent years. During \emph{training} phase of these algorithms, a joint action-value function, including all agents' individual action-value functions, is learned using additional information such as global states, actions, or rewards. Each agent can act independently based on its local observation and its individual action-value function without any additional information in the \emph{execution} phase. Although value-decomposed MARL algorithms have achieved promising success, improving sample efficiency is still a critical problem for MARL. Episodic memory~(EM), which can record the best experiences, has been applied in single-agent RL to improve sample efficiency~\cite{DBLP:conf/icml/PritzelUSBVHWB17,DBLP:journals/corr/BlundellUPLRLRW16,DBLP:conf/nips/LengyelD07,DBLP:conf/ijcai/LinZYZ18}. However, to our best knowledge, EM has not been introduced into MARL. 

In this paper, we make the first attempt to introduce EM into MARL and propose a novel method, called \underline{s}tate-based \underline{e}pisodic \underline{m}emory~(SEM), to improve sample efficiency for MARL. The contributions of this work are briefly outlined as follows:
\begin{itemize}
\item SEM is the first work to introduce EM into MARL to improve sample efficiency. 
\item SEM establishes only one lookup table to record global states and their corresponding highest discounted return among all executed joint actions. SEM replays the highest discounted return from the table as the EM target to supervise the centralized training in the paradigm of CTDE.
\item When using for MARL, SEM can be theoretically proved to have lower space complexity and time complexity than state and action based EM~(SAEM), which is originally proposed for single-agent RL~\cite{DBLP:conf/icml/PritzelUSBVHWB17,DBLP:journals/corr/BlundellUPLRLRW16,DBLP:conf/nips/LengyelD07,DBLP:conf/ijcai/LinZYZ18}.
\item Experimental results on StarCraft multi-agent challenge~(SMAC) show that introducing episodic memory into MARL can improve sample efficiency and SEM can reduce storage cost and time cost compared with SAEM. 
\end{itemize}

\section{Notation}
\label{Background}
In this paper, we adopt similar notations as those in~\cite{DBLP:conf/icml/RashidSWFFW18}. More specifically, we describe the fully cooperative multi-agent task as a decentralized partially observable Markov decision process~(\mbox{Dec-POMDP})~\cite{DBLP:series/sbis/OliehoekA16}. A Dec-POMDP is defined by a tuple $G=<S, U, P, r, Z, O, n, \gamma>$. Here, $S\in \mathbb{R}^F$ denotes the global state of the environment and $U$ denotes the action space of each agent $a \in A \equiv \{1, \ldots, n\}$. At each time step, each agent $a$ chooses an action $u^a\in U$, forming a joint action $\mathbf{u} \in \mathbf{U} \equiv U^{n}$. It results in a transition on the environment according to state transition function $P(s'|s,\mathbf{u}):S \times \mathbf{U} \times S \to [0,1]$. All agents receive the same reward according to the same reward function $r(s,\mathbf{u}): S \times \mathbf{U} \rightarrow \mathbb{R}$ and $\gamma \in[0,1)$ is a discount factor.
Dec-POMDPs consider partially observable scenarios in which each agent has individual observation $o \in O$ according to observation function $Z(s,a): S \times A \to O$. Each agent has an observation-action history $\tau^{a} \in \mathcal{T} \equiv(O \times U)^{*}$ and chooses its action based on a stochastic policy $\pi^{a}(u^a|\tau^{a}): \mathcal{T} \times U \to [0,1]$. $\boldsymbol{\tau} \in \mathbf{T} \equiv \mathcal{T}^{n}$ is the joint action-observation history. The joint action-value function of the joint policy $\pi$ is defined as: $Q_{tot}^{\pi}\left(\boldsymbol{\tau}_{t}, \mathbf{u}_{t}\right)=\mathbb{E}_{\boldsymbol{\tau}_{t+1: \infty}, \mathbf{u}_{t+1: \infty}}\left[R_{t} \mid \boldsymbol{\tau}_t, \mathbf{u}_{t}\right]$, where $R_{t}=\sum_{i=0}^{\infty} \gamma^{i} r_{t+i}$ is the discounted return.

\section{Related Work}
\subsection{EM for Single-Agent RL}
In single-agent RL, deep reinforcement learning algorithms need to take millions of interactions with the environments to attain human-level performance. However, humans can quickly exploit the high reward after the first discovery by using the hippocampus, which can record EM. Motivated by the hippocampus' ability, researchers proposed to use EM to achieve fast learning and improve sample efficiency for single-agent RL. Model-free episodic control~(MFEC)~\cite{DBLP:journals/corr/BlundellUPLRLRW16} and neural episodic control~(NEC)~\cite{DBLP:conf/icml/PritzelUSBVHWB17} try to use lookup tables to record the EM and retrieve useful values from lookup tables for action selection. MFEC and NEC can be seen as tabular RL methods, which lack good generalization compared with deep neural network-based RL methods. Episodic memory deep q-network~(EMDQN)~\cite{DBLP:conf/ijcai/LinZYZ18} combines EM with deep q-network~(DQN) to achieve good generalization and improve sample efficiency by accelerating the training process of DQN. EM has been widely applied in single-agent RL, but it has not been introduced into MARL. Furthermore, all existing EM-based single-agent RL methods adopt state and action based EM~(SAEM). SAEM will face many difficulties when SAEM is applied into MARL, which will be detailedly described in Section~\ref{sec:SAEM}.

\subsection{Value-based MARL Algorithms}
The most common and straightforward value-based approach in the multi-agent setting is to break down a multi-agent learning problem into multiple independent single-agent learning problems, which is called decentralized value-based methods. One of the representative methods is independent Q-learning~(IQL)~\cite{tan1993multi}. Decentralized value-based methods benefit from scalability because each agent can make decisions based on a decentralized policy. However, each agent has to treat other agents as a part of the environment. The other agents' policies will change during the training procedure, making the environment not stationary. Therefore, these decentralized value-based methods suffer from sample inefficiency and instability problems.  

The sample inefficiency and instability problems caused by decentralized policies can be alleviated by adding centralized training into MARL, resulting in a MARL paradigm called centralized training with decentralized execution~(CTDE)~\cite{DBLP:journals/jair/OliehoekSV08}. Benefiting from the paradigm of CTDE, researchers have proposed some approaches, called value-decomposed MARL algorithms, that learn a centralized but decomposed Q value function to improve agent learning performance in recent years~\cite{DBLP:conf/atal/SunehagLGCZJLSL18,DBLP:conf/icml/RashidSWFFW18,DBLP:conf/icml/SonKKHY19,DBLP:conf/nips/RashidFPW20,DBLP:journals/corr/abs-2008-01062}. During execution, each agent still acts independently. In the value-decomposed MARL algorithms, individual-global-max~(IGM) is a crucial principle, ensuring that the optimal joint action across agents in the joint action-value~$Q_{t o t}\left(\boldsymbol{\tau}, \mathbf{u};\theta\right)$ and the collection of all optimal individual action in the $[Q_{a}(\tau^{a},u^{a};\theta)]_{a=1}^{n}$ are consistent. The parameters $\theta$ are learnred by minimizing the following expected TD error:
\begin{equation}
	L(\theta)=\sum_{b=1}^{B}\sum_{t=1}^{T}(Q_{t o t}\left(\boldsymbol{\tau}_t^b, \mathbf{u}_t^b;\theta\right)-y_t^b)^{2},
\label{tderror}
\end{equation}
where $y_t^b = r_t^b + \gamma \max_{\mathbf{u}} Q_{tot}(\boldsymbol{\tau}_{t+1}^{b},\mathbf{u};\theta^{-})$ and $\theta^{-}$ are the parameters of a target network that are periodically copied from $\theta$. 
VDN~\cite{DBLP:conf/atal/SunehagLGCZJLSL18} and QMIX~\cite{DBLP:conf/icml/RashidSWFFW18} respectively propose two decomposition structures, additivity and monotonicity, which are sufficient conditions for the IGM. WQMIX~\cite{DBLP:conf/nips/RashidFPW20} tries to use a weighted projection that places more importance on better joint actions to overcome the limitation of QMIX. QTRAN~\cite{DBLP:conf/icml/SonKKHY19} tries to realize the entire IGM function class by using extra soft regularizations, which actually loses the IGM guarantee. QPLEX~\cite{DBLP:journals/corr/abs-2008-01062} uses a duplex dueling architecture and provides a guaranteed IGM consistency. Although these value-decomposed MARL algorithms can outperform decentralized value-based methods, they still suffer from sample inefficiency and instability problems.

\section{State and Action based Episodic Memory for MARL}\label{sec:SAEM}
The EM for single-agent RL~\cite{DBLP:conf/icml/PritzelUSBVHWB17,DBLP:journals/corr/BlundellUPLRLRW16,DBLP:conf/nips/LengyelD07,DBLP:conf/ijcai/LinZYZ18} is defined on state and action. This is why we call it state and action based EM~(SAEM) in this paper. To the best of our knowledge, EM, including SAEM, has not been introduced into MARL before. In this section, we will extend the application of SAEM from single-agent RL to MARL, by replacing the action in single-agent RL with the joint action in MARL.

More specifically, we establish a lookup table for each joint action and denote this lookup table as $Q^{\text{SA}}(s,\mathbf{u})$, where $s$ is the global state and $\mathbf{u}$ is the joint action. Each entry in the table $Q^{\text{SA}}(s,\mathbf{u})$ records the highest return ever obtained by taking joint action $\mathbf{u}$ in state $s$. At the end of the episode, we store the episode $(\mathbf{o}_1, \mathbf{u}_1, s_1,r_1, ...,\mathbf{o}_T, \mathbf{u}_T, s_T,r_T)$ into the replay buffer $H$ and store $(s_t,\mathbf{u}_t, R(s_t))$ into a set $M$ of size $|M|$, where $R(s_t)=\sum_{i=t}^{T} \gamma^{i-t} r_{i}$ is the discounted return received after taking joint action $\mathbf{u}_{t}$ in state $s_t$. $Q^{\text{SA}}$ is updated when the set $M$ is filled:
\begin{equation}
	Q^{\text{SA}}\left(s,\mathbf{u}\right) \leftarrow\left\{\begin{array}{l}R_{\text{Max}}(s,\mathbf{u}),~~~~~~~~~~~~~~~~~~\text{if }(s,\mathbf{u})\notin Q^{\text{SA}}; \\ \max \left\{Q^{\text{SA}}(s,\mathbf{u}), R_{\text{Max}}(s,\mathbf{u})\right\},~~~\text{otherwise},\end{array}\right.
\label{eq:updateem}
\end{equation}
where $R_{\text{Max}}(s,\mathbf{u}) = \max_k \{R_k(s,\mathbf{u})\}$ and $(s,R_k(s,\mathbf{u}))\in M$, $k \in\{1,2,...,K\}$. For any state and joint action, we update their values with the highest discounted return. After updating $Q^{\text{SA}}(s,\mathbf{u})$, the set $M$ is made empty. The value of each entry in the table $Q^{\text{SA}}$ is updated increasingly and the number of entries also increases during training. We limit the maximum size of the table $Q^{\text{SA}}$ and remove the least frequently assessed entry when $Q^{\text{SA}}$ is filled.

In the centralized training phase, the loss function of SAEM is defined as:
\begin{equation}
\label{multiem}
\begin{aligned}
L(\theta)=\sum_{b=1}^{B}\sum_{t=1}^{T} &(1-\lambda) \left( Q_{tot} \left(\boldsymbol{\tau}_t^b,\mathbf{u}_t^b;\theta \right)-y_t^b \right)^{2}+\lambda \left(Q_{tot} \left(\boldsymbol{\tau}_t^b,\mathbf{u}_t^b;\theta \right)-E_{s_t}^{b,\mathbf{u}_t^b}\right)^{2},
\end{aligned}
\end{equation}
where $y_t^b = r_t^b + \gamma \max_{\mathbf{u}} Q_{tot}(\boldsymbol{\tau}_{t+1}^b,\mathbf{u};\theta^{-})$ is the vanilla target in the value-decomposed MARL algorithms and $E_{s_t}^{b,\mathbf{u}_t^b}=Q^{\text{SA}}(s_{t}^b,\mathbf{u}_t^b)$ is the episodic memory target recorded by $Q^{\text{SA}}$. $\lambda\in [0,1]$ is the coefficient to balance the trade-off between the two targets. 

We can see that SAEM suffers from two deficiencies when using for MARL:  high space complexity and high time complexity. Because SAEM needs $|U|^n$ lookup tables, the space complexity is $\mathcal{O}(c|U|^n)$, where $c>1$ is a constant. Hence, the space complexity is exponentially higher than that in single-agent RL. The time complexity is also high. In the worst case, there are $|M|$ different joint actions, and hence $|M|$ tables are needed to be updated when using the set $M$ to update $Q^{\text{SA}}$. Hence, the time complexity of updating $Q^{\text{SA}}$ is $\mathcal{O}(|M|)$ in the worse case. This motivates us to design new EM mechanisms like state-based EM in the next section.

\section{State-based Episodic Memory for MARL}
\label{sec:SEM}
This section introduces our newly proposed EM, called state-based episodic memory~(SEM), for MARL.

\subsection{Lookup Table in SEM}
In SEM, we establish only one lookup table $Q^{\text{S}}(s)$, which is indexed by global states. Given the global state $s$, each entry in $Q^{\text{S}}(s)$ records the highest return ever obtained. At the end of episode, we store the episode $(\mathbf{o}_1, \mathbf{u}_1, s_1,r_1, ...,\mathbf{o}_T, \mathbf{u}_T, s_T,r_T)$ into the replay buffer $H$ and store $(s_t, R(s_t))$ into a set $M$, where $R(s_t)=\sum_{i=t}^{T} \gamma^{i-t} r_{i}$ is the discounted return received after taking joint action $\mathbf{u}_{t}$ in state $s_t$. $Q^{\text{S}}$ is updated according to the set $M$:
\begin{equation}
	Q^{\text{S}}\left(s_{t}\right) \leftarrow\left\{\begin{array}{l}R_{\text{Max}}(s_t),~~~\text{if }s_{t}\notin Q^{\text{S}}; \\ \max \left\{Q^{\text{S}}\left(s_{t}\right), R_{\text{Max}}(s_t)\right\},~~~\text{otherwise},\end{array}\right.
\label{eq:updateem}
\end{equation}
where $R_{\text{Max}}(s_t) = \max_k \{R_k(s_t)\}$ and $(s_t,R_k(s_t))\in M$, $k \in\{1,2,...,K\}$. For any state, we update their values with the highest return. After updating $Q^{\text{S}}$, the set $M$ is made empty. The value of each entry in the table $Q^{\text{S}}$ is updated increasingly and the number of entries also increases during training. We limit the maximum size of the table $Q^{\text{S}}$ and remove the least frequently assessed entry when $Q^{\text{S}}$ is filled.

\subsection{Training Procedure}
In the centralized training procedure, the loss function of \mbox{SEM} is defined as:
\begin{equation}
\begin{aligned}
L(\theta)=\sum_{b=1}^{B}\sum_{t=1}^{T} &(1-\lambda) (Q_{tot}(\boldsymbol{\tau}_t^b,\mathbf{u}_t^b;\theta)-y_t^b)^{2}+\lambda (Q_{tot}(\boldsymbol{\tau}_t^b,\mathbf{u}_t^b;\theta)-E_t^b)^{2},
\end{aligned}
\label{EM_loss}
\end{equation}
where $y_t^b = r_t^b + \gamma \max_{\mathbf{u}} Q_{tot}(\boldsymbol{\tau}_{t+1}^b,\mathbf{u};\theta^{-})$ and $E_t^b= r_t^b + \gamma Q^{\text{S}}(s_{t+1}^b)$. $\theta^{-}$ is the parameter of the target network which is copied from $\theta$. $y_t^b$ is a target inferred by the target network and $E_{t}^{b}$ is an episodic memory target. $\lambda\in [0,1]$ is the coefficient to balance the trade-off between two targets. When $\lambda$ is set to 0, our method degenerates to the original value-decomposed MARL algorithm. 

\begin{figure*}[tb]
\centering
\vspace{-0.4cm}
\includegraphics[width=0.9\linewidth]{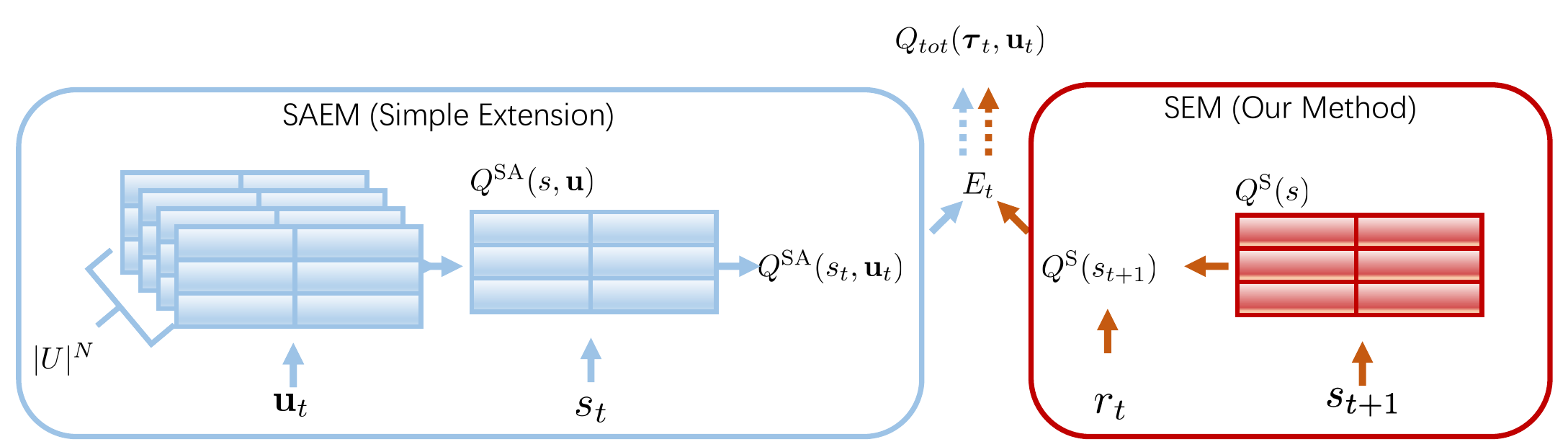}
\caption{Comparison between SAEM architecture for MARL and SEM architecture. Best viewed in color.}
\label{fig:architecture}	
\end{figure*}

During training, the vanilla target $y_t^b$, approximated by using the target network, might be overestimated, which could mislead the training. The episodic memory target $E_t^b$~(or $E_s^{b,\mathbf{u}}$ in SAEM) is more stable than the vanilla target, because the episodic memory target is retrieved from the lookup table, rather than approximated by the function. Furthermore, the maximum operator in updating $Q^{\text{S}}(s)$ not only records the highest return but also plays the role of the maximum operator $\max_{\mathbf{u}} Q_{tot}(\boldsymbol{\tau}_{t+1}^b,\mathbf{u};\theta^{-})$, because we ignore the action $\mathbf{u}$ given the state $s$ in the $Q^{\text{S}}(s)$. If we cannot find the $Q^{\text{S}}(s_{t+1}^{b})$ in the lookup table, we use $\max_{\mathbf{u}} Q_{tot}(\boldsymbol{\tau}_{t+1}^b,\mathbf{u};\theta^{-})$ to replace $Q^{\text{S}}(s_{t+1}^{b})$. Algorithm~\ref{alg:em} briefly presents the training procedure of \mbox{SEM}. The architecture of SEM is shown in Figure~\ref{fig:architecture}. Please note that each agent takes actions based on $Q_{a}$ and our method can be combined with all existing value-decomposed MARL algorithms.


\subsection{Complexity}\label{sec:complexity}
SEM only need one table to store and update, rather than $|U|^n$ tables in SAEM. The space complexity of SEM is $O(c_{s})$, where $c_{s}$ is a constant. When $Q^{\text{S}}(s)$ is updated by the set $M$, it needs to update the $Q^{\text{S}}(s)$ only once. Hence, the time complexity of SEM is $O(1)$. Hence, we can theoretically prove that SEM has lower space complexity and time complexity than SAEM, when using for MARL.
\subsection{State Representation}
Although SEM has lower space complexity than SAEM, the storage cost of all global states in the table $Q^{\text{S}}(s)$ might still be large if the dimension of state space is high. Similar to~\cite{DBLP:journals/corr/BlundellUPLRLRW16,DBLP:conf/icml/PritzelUSBVHWB17,DBLP:conf/ijcai/LinZYZ18}, we utilize random projection $\phi$ to project a state from the original global state space $S$ with dimension $F$ into a lower-dimensional space with dimension $D$. The random projection $\phi$ is denoted as $\phi(s):s \to Vs$, where each entry in $V \in \mathbb{R}^{D \times F}$ is randomly drawn from a standard Gaussian. Based on Johnson-Lindenstrauss lemma~\cite{johnson1984extensions}, when the random matrix $V$ is drawn from a standard Gaussian, this transformation approximately keeps relative distances in the original space. Projecting global states to lower-dimension vectors can accelerate the speed of table lookup. Please note that we still use $Q^{\text{S}}(s)$ to denote the table in SEM, omitting $\phi$ although random projection is adopted in this paper. This state representation can also be adopted in SAEM.

\begin{algorithm}[tb]
\caption{State-based Episodic Memory~\mbox{(SEM)} for MARL}
\label{alg:em}
\begin{algorithmic} 
\STATE Initialize a replay buffer $H$, an empty set $M$, an episodic memory table $Q^{\text{S}}$.
\STATE Initialize network parameter $\theta$ and $\theta^{-}= \theta$.
\FOR{each episode}
\FOR{$t=1,2,3,...,T$}
\STATE Receive observation $[o_t^a]_{a=1}^{n}$, global state $s_t$.
\STATE Select a random action $u_t^a$ with probability $\epsilon$, otherwise $u_t^a=\arg \max_{u^a}Q_a(\tau^a, u^a)$ for each agent $a$.
\STATE Take action $[u_t^a]_{a=1}^{n}$. 
\ENDFOR
\STATE Store the episode $(\mathbf{o}_1, \mathbf{u}_1, s_1,r_1...,\mathbf{o}_T, \mathbf{u}_T, s_T, r_T)$ in $H$.
\FOR{$t=T,T-1,...,1$}
\STATE Let $s_t=\phi(s_t)$. 
\STATE Compute $R_t$ and store $(s_t, R_t)$ into $M$.
\ENDFOR 
\STATE Sample $B$ episodes from the replay buffer $H$.
\STATE Compute $y_t^b$ and $E_t^b$.
\STATE Update $\theta$ by minimizing the loss in~(\ref{EM_loss}).
\STATE Update target network parameter $\theta^{-} = \theta$ with period $I_n$. 
\STATE Update $Q^{\text{S}}(s_t)$ using $M$ according to~(\ref{eq:updateem}) when $M$ is filled and then empty $M$.
\ENDFOR
\end{algorithmic}
\end{algorithm}




\section{Experiment}
\label{Experiments}
\subsection{StarCraft Multi-Agent Challenge}
\label{SMAC}
StarCraft multi-agent challenge~(SMAC)~\footnote{We use $\mathrm{SC} 2.4 .6 .2 .69232$~(the same version as that in~\cite{DBLP:conf/atal/SamvelyanRWFNRH19}), instead of the newer version $\mathrm{SC} 2.4.10$.}, based on the real-time strategy game StarCraft \uppercase\expandafter{\romannumeral2} and the SC2LE environment~\cite{DBLP:journals/corr/abs-1708-04782}, is a popular benchmark for cooperative multi-agent RL. SMAC focuses on micromanagement challenges. In SMAC, it involves two armies, one controlled by the build-in AI and the other controlled by the user. Each unit can be controlled by an independent agent. At each time-step, agents receive their local observations, which depend on their sight range. The agents are allowed to take actions, including move[direction], attack[enemy\_id], stop and no-op. Agents can only move in four directions: north, south, east, or west. If the enemy is within the agent's shooting range, the agent can perform the action attack[enemy\_id]. The maximum number of actions an agent can take ranges between 7 and 70, depending on the scenario. The goal of agents is to maximize the win rate for each battle scenario. The default setting of SMAC is to use the shaped reward. 
For evaluation, we run 32 evaluation episodes without any exploratory behaviors every 10000 time-steps. More details about the implementation are included in Appendix of the supplementary materials. 

\subsection{Effectiveness of Episodic Memory}
\begin{wrapfigure}{r}{6cm}
\vspace{-1.4cm}
\includegraphics [width=5.8cm,clip]{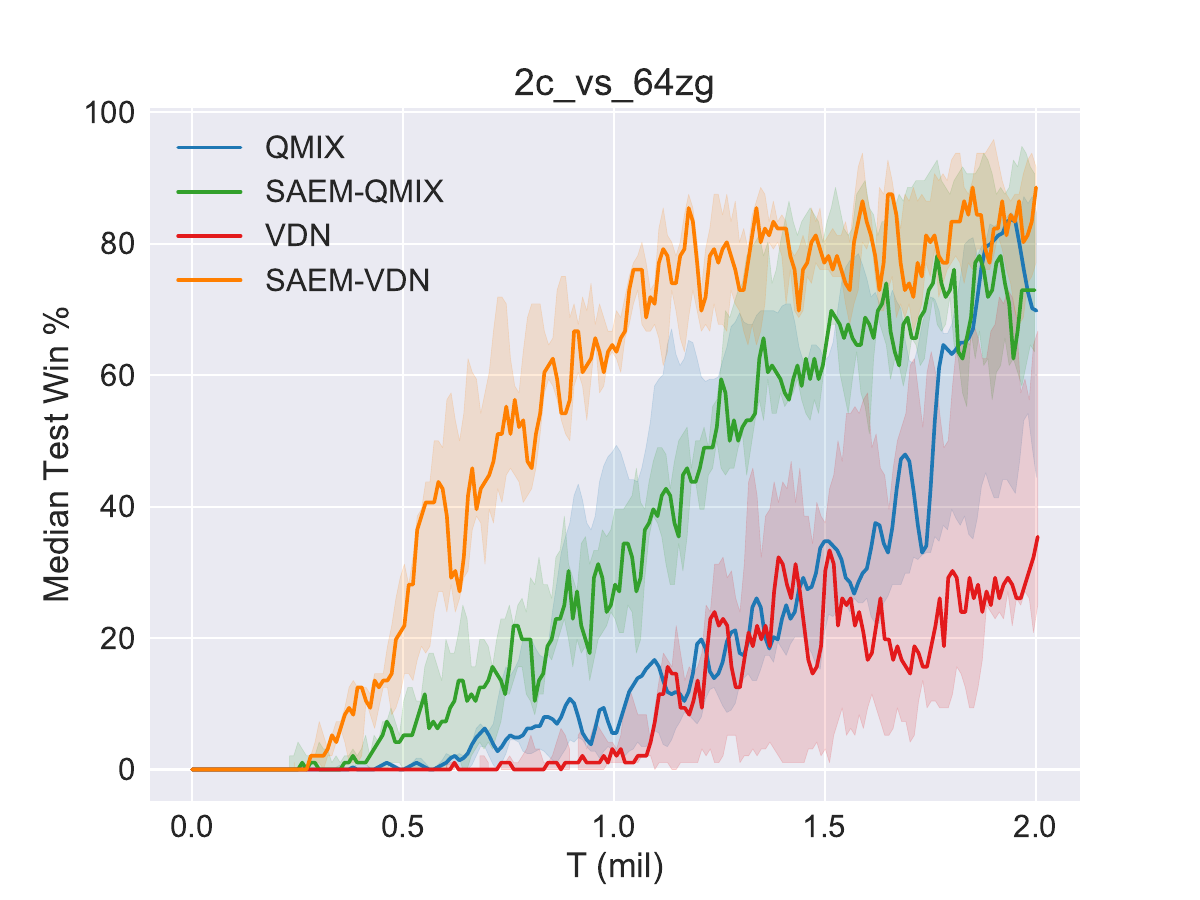}
\caption{Results of SAEM-VDN, SAEM-QMIX and baselines~(VDN and QMIX) on 2c\_vs\_64zg, including median test win rate as well as the 25-75\% percentiles.}
\label{result_SAEM}
\end{wrapfigure}
We evaluate SAEM on 2c\_vs\_64zg, a hard map on SMAC, to verify the effectiveness of EM for MARL. We choose two classic value-decomposed MARL algorithms, VDN and QMIX, as the baselines. We combine SAEM with VDN and QMIX, which are denoted as SAEM-VDN and SAEM-QMIX, respectively. The coefficient $\lambda$ is set to 0.1. The size of the lookup table is set to 1 million and the size of $M$ is set to 5K. Other hyper-parameters are described in Appendix. The results are shown in Figure~\ref{result_SAEM}. We can find that the methods with SAEM have better performance than baselines without episodic memory, verifying that introducing episodic memory into the multi-agent setting is effective.

\subsection{Results of SEM}
\label{Results of SEM}
We evaluate our method, SEM, on the eight maps of SMAC. These maps include 1c3s5z, 2s\_vs\_1sc, 2s3z, 3s5z, 27m\_vs\_30m, 2c\_vs\_64zg, MMM2 and bane\_vs\_bane. The snapshots and configurations of these maps are shown in the Appendix. We choose several value-decomposed MARL algorithms as baselines, including VDN~\cite{DBLP:conf/atal/SunehagLGCZJLSL18}, QMIX~\cite{DBLP:conf/icml/RashidSWFFW18}, QPLEX~\cite{DBLP:journals/corr/abs-2008-01062}, WQMIX~\cite{DBLP:conf/nips/RashidFPW20}~\footnote{The weighting function of WQMIX is the centrally-weighting function.}. The coefficient $\lambda$ is set to $0.1$. The size of the lookup table is set to 1 million and the size of set $M$ is set to 5K, which means that the lookup table is updated every 5K time-steps. Other hyper-parameters are described in Appendix. We show our results, mean and median scores\footnote{The formula of mean scores and the formula of median scores are shown in the Appendix.} over eight maps of SMAC, at 0.25M time-steps and 0.5M time-steps in Table~\ref{tb_results}. We can find that the mean scores and median scores of SEM combined with value-decomposed MARL algorithms all surpass those of the corresponding vanilla value-decomposed MARL algorithms. It verifies that SEM can improve sample efficiency compared with baselines that do not use episodic memory. 

\begin{figure*}[tb]
\centering
\vspace{-0.4cm}
\subfigure[1c3s5z]{
\label{fig:1c3s5z} 
\includegraphics[width=0.34\linewidth,height=0.23\linewidth]{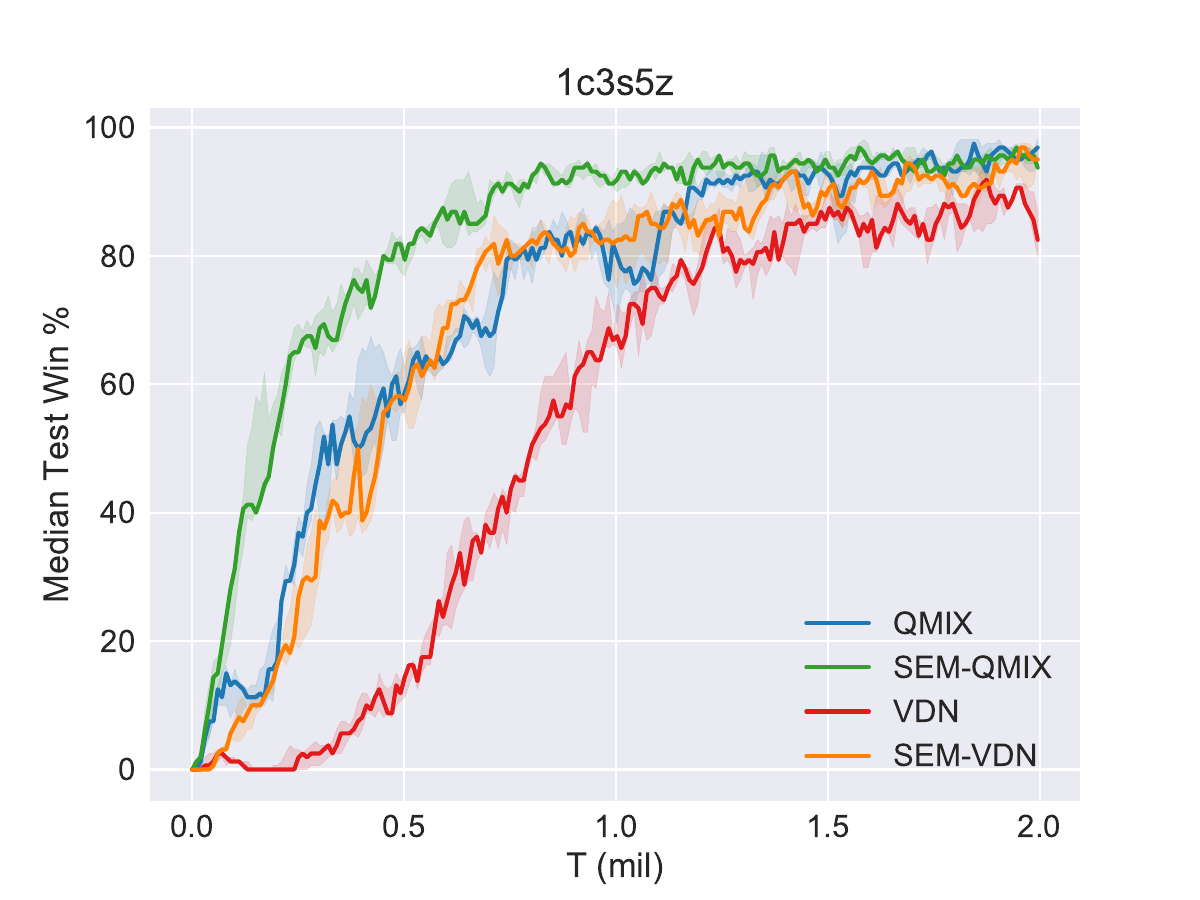}}
\hspace{-0.05\linewidth}
\subfigure[2s3z]{
\label{fig:2s3z} 
\includegraphics[width=0.34\linewidth,height=0.23\linewidth]{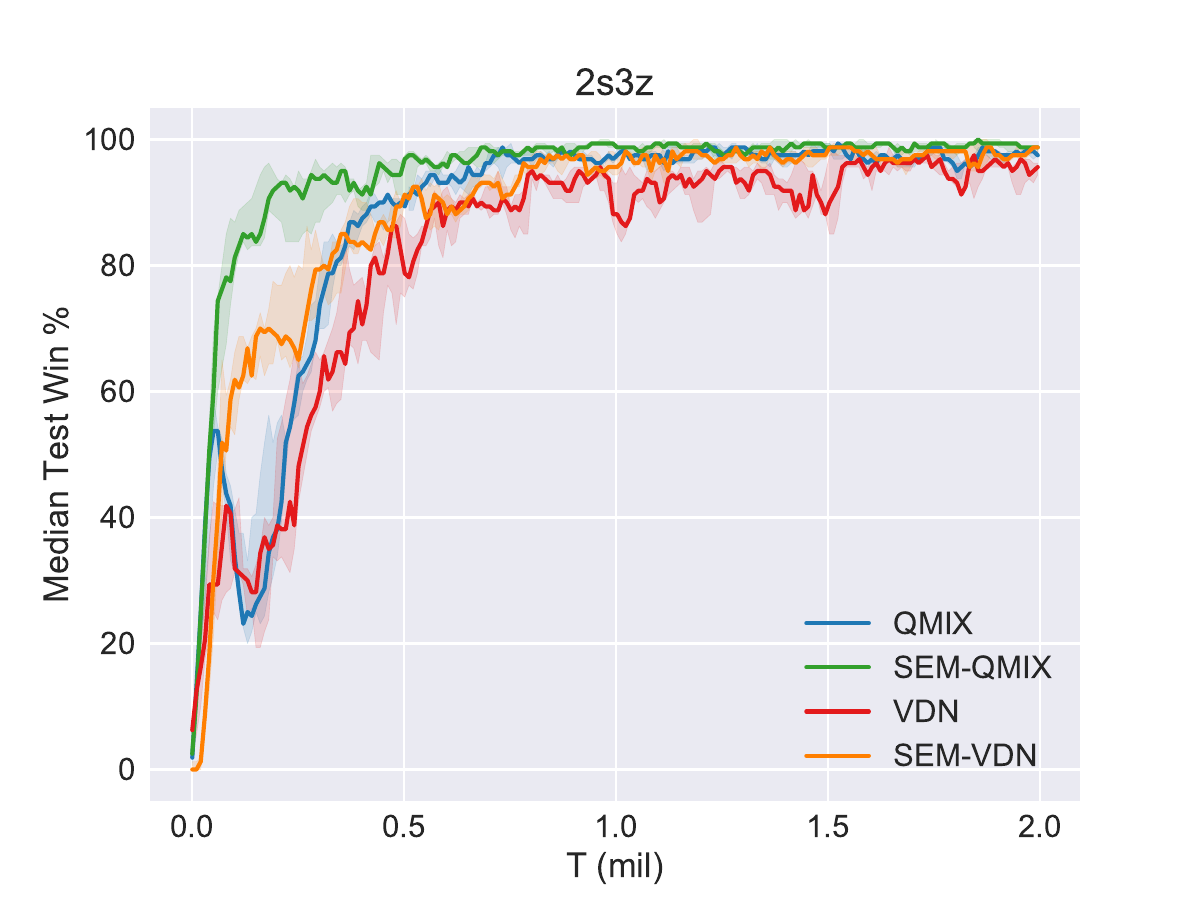}}
\hspace{-0.05\linewidth}
\subfigure[3s5z]{
\label{fig:3s5z} 
\includegraphics[width=0.34\linewidth,height=0.23\linewidth]{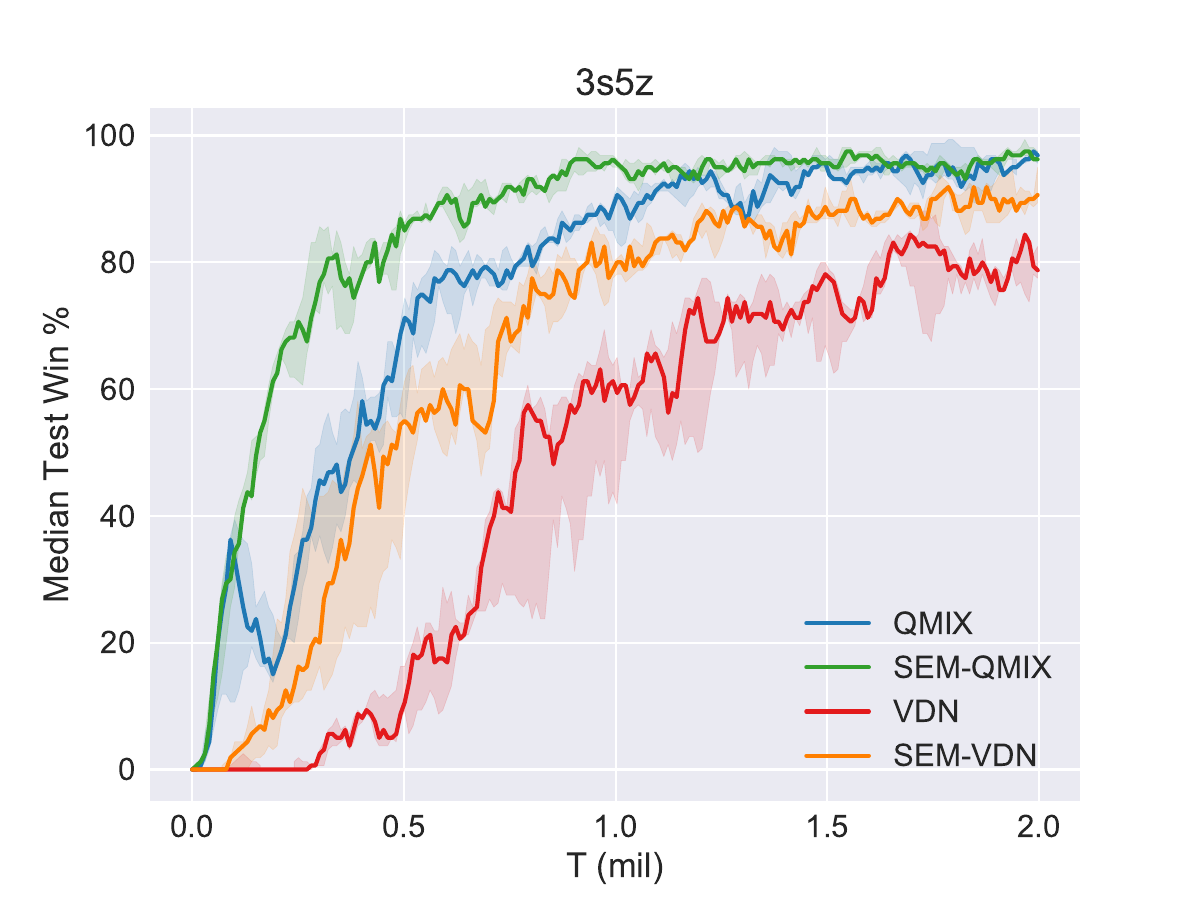}}\\
\hspace{-0\linewidth}
\vspace{-0.3cm}
\subfigure[27m\_vs\_30m]{
\label{fig:27m30m} 
\includegraphics[width=0.34\linewidth,height=0.23\linewidth]{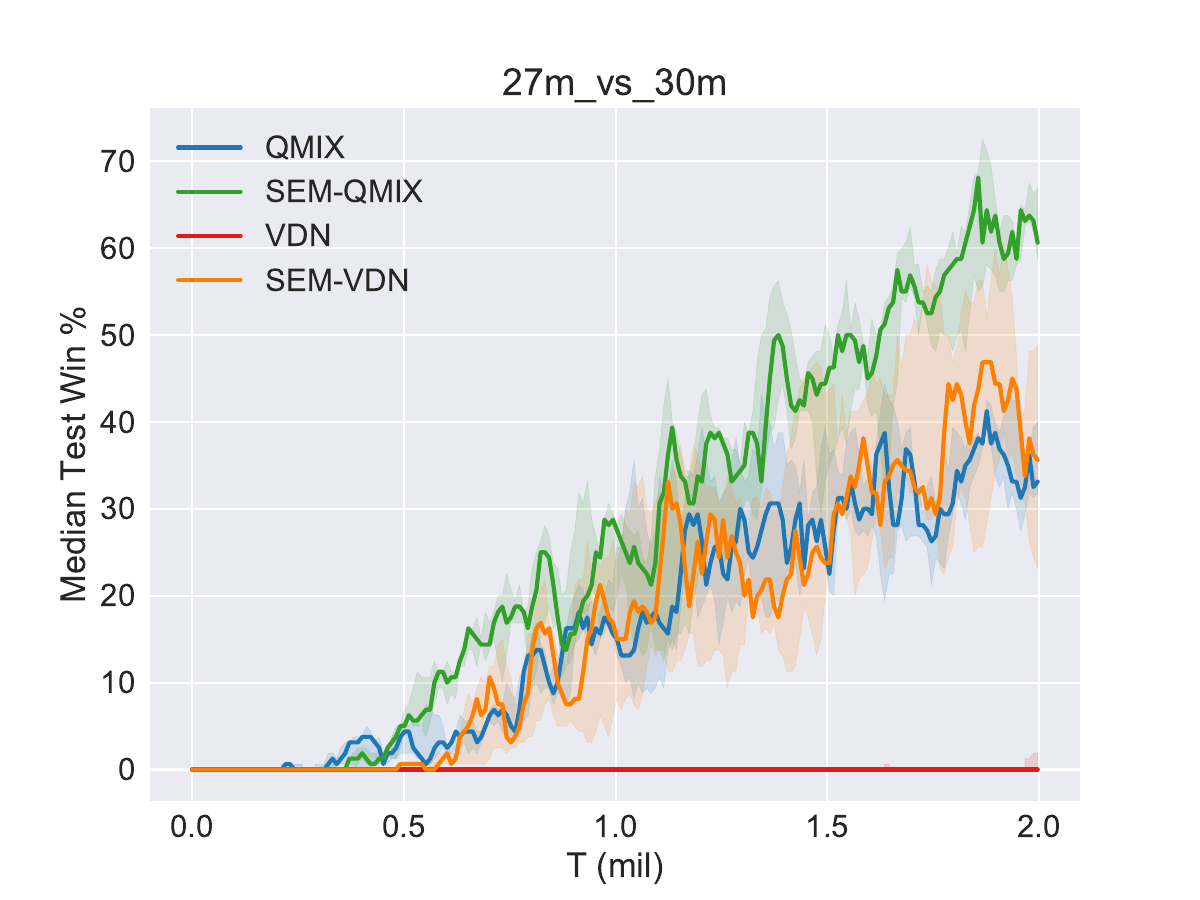}}
\hspace{-0.05\linewidth}
\subfigure[2c\_vs\_64zg]{
\label{fig:2c64zg} 
\includegraphics[width=0.34\linewidth,height=0.23\linewidth]{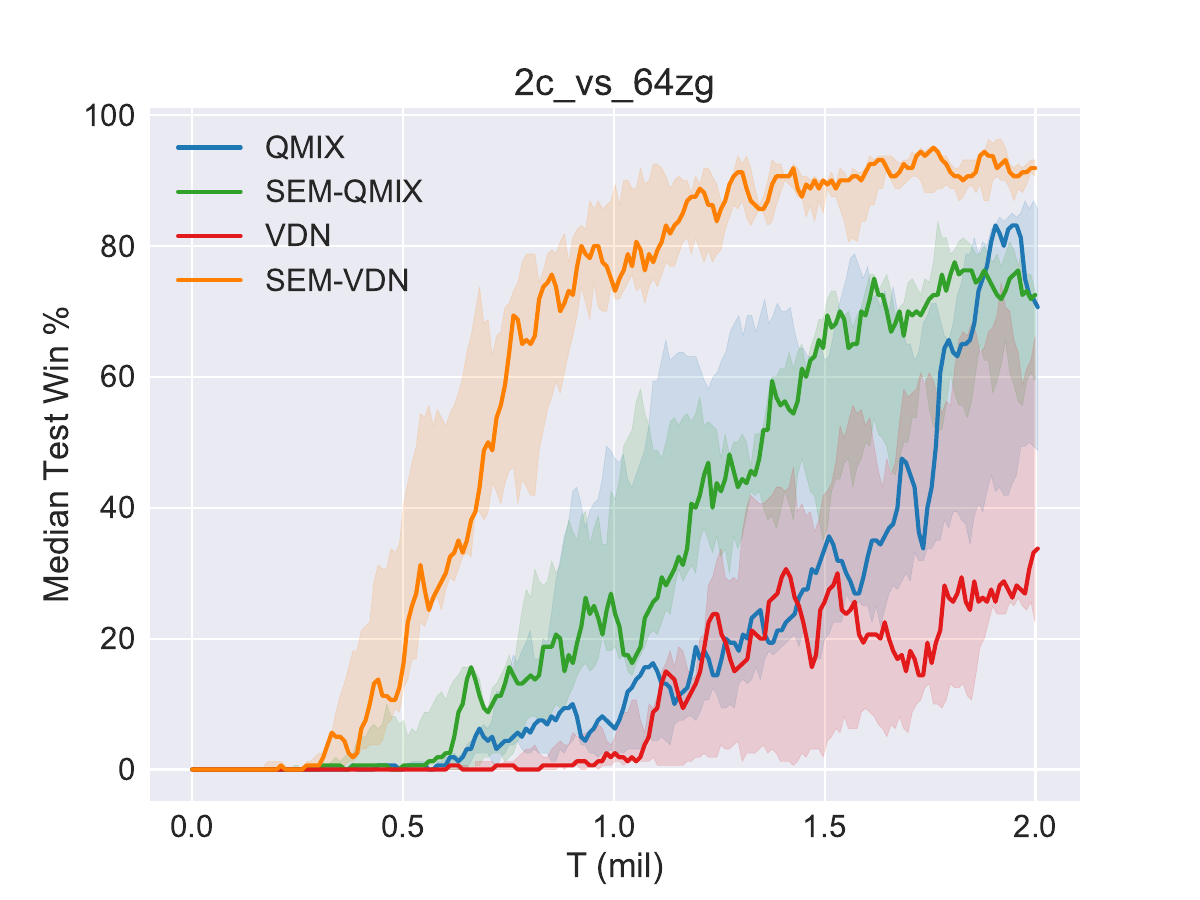}}
\hspace{-0.05\linewidth}
\subfigure[bane\_vs\_bane]{
\label{fig:banebane} 
\includegraphics[width=0.34\linewidth,height=0.23\linewidth]{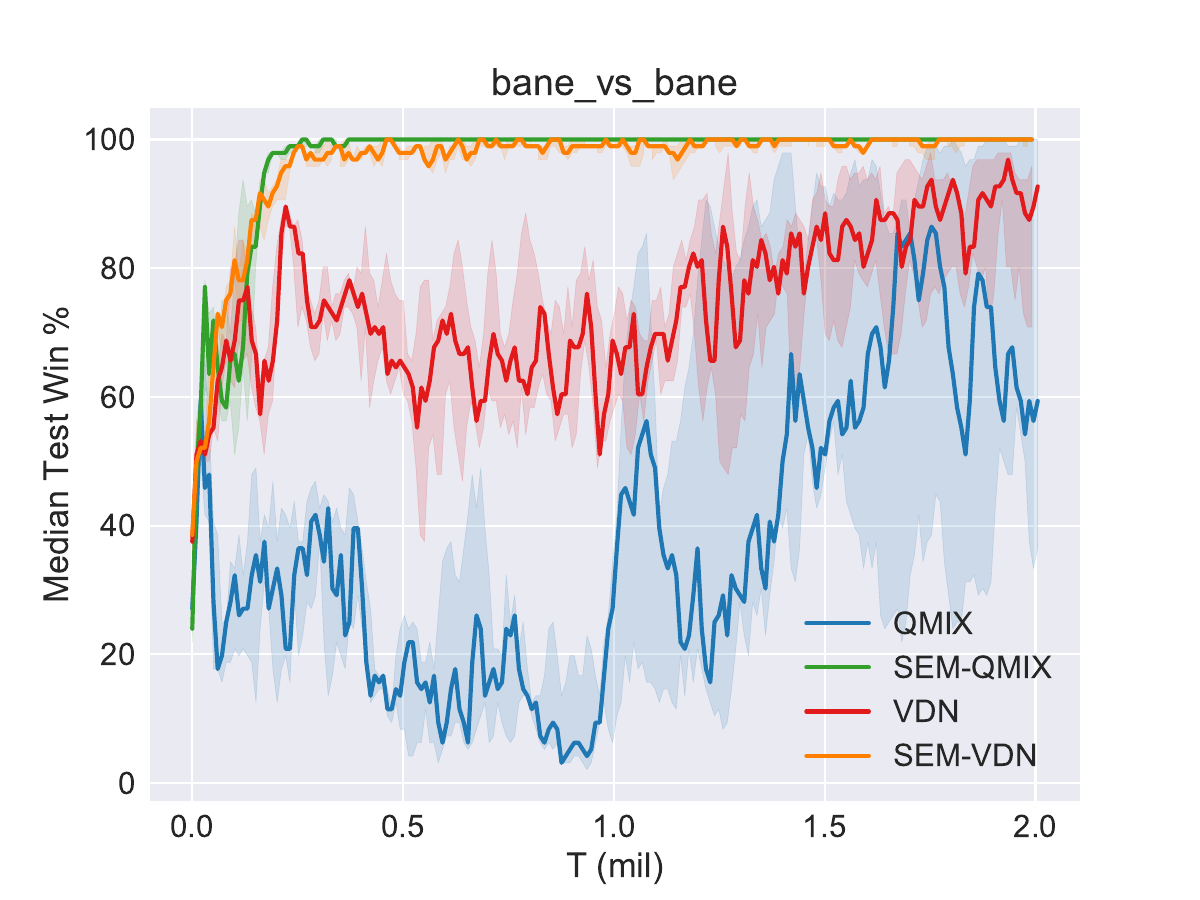}}
\caption{Results of our methods~(SEM-VDN and SEM-QMIX) and baselines~(VDN and QMIX), including the median performance as well as the 25-75\% percentiles.  }
\vspace{-0.2cm}
\label{EMQMIX} 
\end{figure*}

In Figure~\ref{EMQMIX}, we show the training curves of SEM-VDN, SEM-QMIX, VDN and QMIX on six maps\footnote{Due to the limitation of pages, the training curves of SEM-VDN, SEM-QMIX, VDN, and QMIX on the other two maps are shown in the Appendix.}. Our methods, \mbox{SEM}-QMIX and \mbox{SEM}-VDN, can converge faster than corresponding baselines and improve sample efficiency on all maps. On 2s3z, the performances of QMIX and VDN both drop sharply at about $0.2$ million time-steps. Our method can alleviate this phenomenon obviously since episodic memory could help the agents to remember the best experience. On 27m\_vs\_30m, \mbox{SEM-VDN} can achieve $40\%$ test battle won while VDN is bound to fail in all the battles. The performance of \mbox{SEM}-QMIX has also increased by about $15\%$ compared to that of QMIX. On 2c\_vs\_64zg, \mbox{SEM-VDN} can achieve $80\%$ test battle won while VDN fails in all battles. \mbox{SEM}-QMIX can learn faster than QMIX. For bane\_vs\_bane, the results of QMIX and VDN exhibit a large variance. 
\begin{table}[htb]
\caption{Mean and median scores for eight maps on SMAC at 0.25M time steps and 0.5M time steps. "w/o" represents the vanilla baselines and "w/" represents the SEM combined with the baselines. Boldface numbers indicate best results. The results of each map is shown in the Appendix.}
\centering
\begin{tabular}{l|p{0.06\linewidth}<{\centering}p{0.06\linewidth}<{\centering}|p{0.06\linewidth}<{\centering}p{0.06\linewidth}<{\centering}||p{0.06\linewidth}<{\centering}p{0.06\linewidth}<{\centering}|p{0.06\linewidth}<{\centering}p{0.06\linewidth}<{\centering}}
\toprule
       & \multicolumn{4}{c||}{0.25M}                                          & \multicolumn{4}{c}{0.5M}                                          \\
       \cline{2-9}
Baselines       & \multicolumn{2}{c|}{Mean} & \multicolumn{2}{c||}{Median} & \multicolumn{2}{c|}{Mean} & \multicolumn{2}{c}{Median} \\
       \cline{2-9}
       & w/o         & w/         & w/o          & w/           & w/o         & w/          & w/o          & w/            \\
       \hline
VDN    & 26\%         & \textbf{36\%}       & 1\%           & \textbf{25\%}          & 33\%          & \textbf{55\%}        & 14\%        & \textbf{64\%}         \\
QMIX   & 21\%          & \textbf{44\%}         & 18\%           & \textbf{47\%}          & 35\%          & \textbf{55\%}         & 28\%           & \textbf{76\%}            \\
QPLEX  & 45\%          & \textbf{50\%}         & 46\%           & \textbf{69\%}          & 58\%          & \textbf{63\%}         & 84\%          & \textbf{90\%}            \\
WQMIX & 26\%          & \textbf{42\%}         & 20\%           & \textbf{40\%}          & 46\%          & \textbf{52\%}         & 49\%           & \textbf{54\%}            \\
\bottomrule             
\end{tabular}
\label{tb_results}
\end{table}
Our methods, \mbox{SEM}-QMIX and \mbox{SEM}-VDN, can both outperform the baselines (QMIX and VDN) with a large margin and the median test win of our methods converges to 1 quickly.

\subsection{Comparison among Different Targets}
\begin{figure*}[htb]
\centering
\vspace{-0.1cm}
\subfigure[VDN]{
\label{fig:vdn2c64zgtarget}
\includegraphics[width=0.3\linewidth]{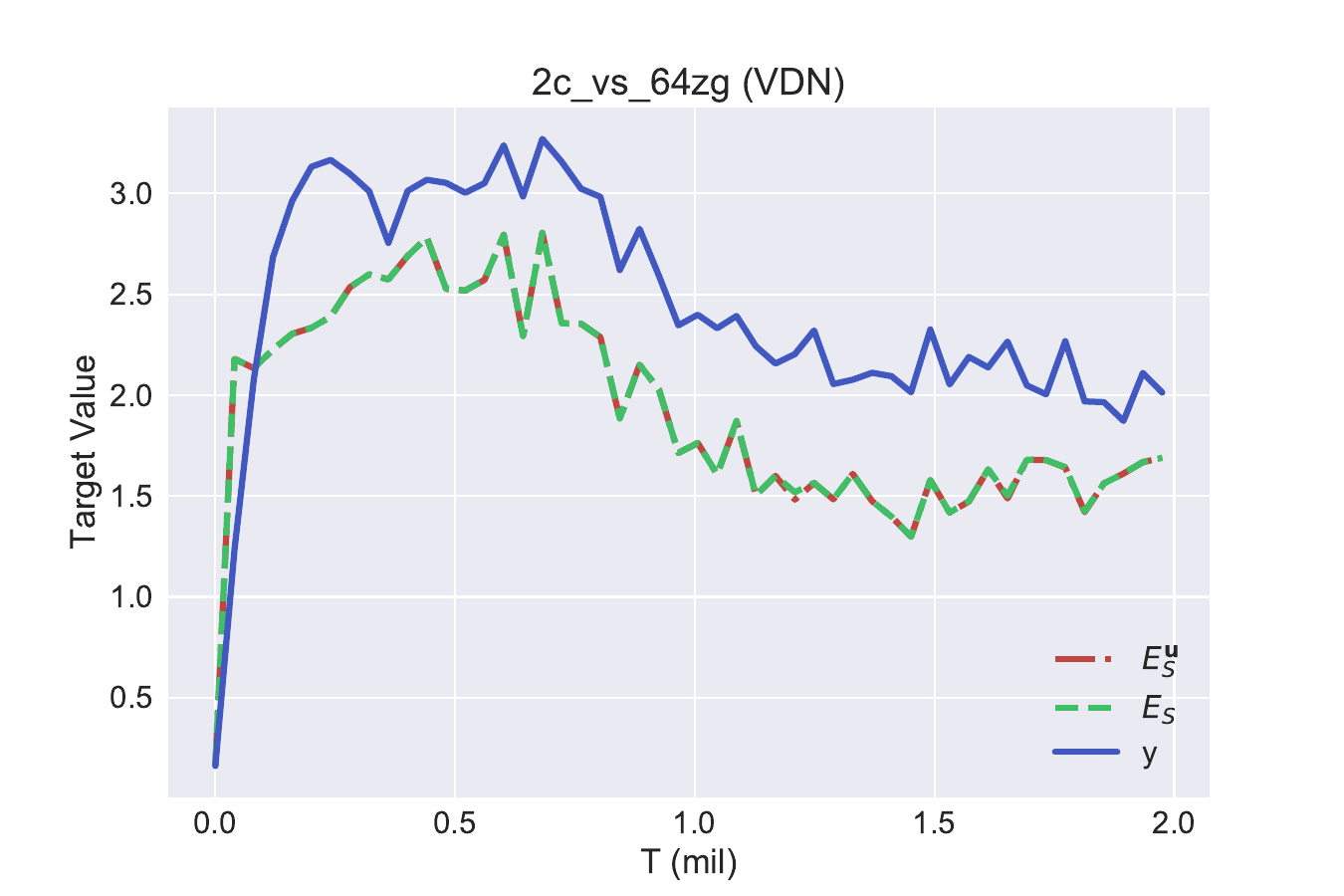}}
\subfigure[SEM-VDN]{
\includegraphics[width=0.3\linewidth]{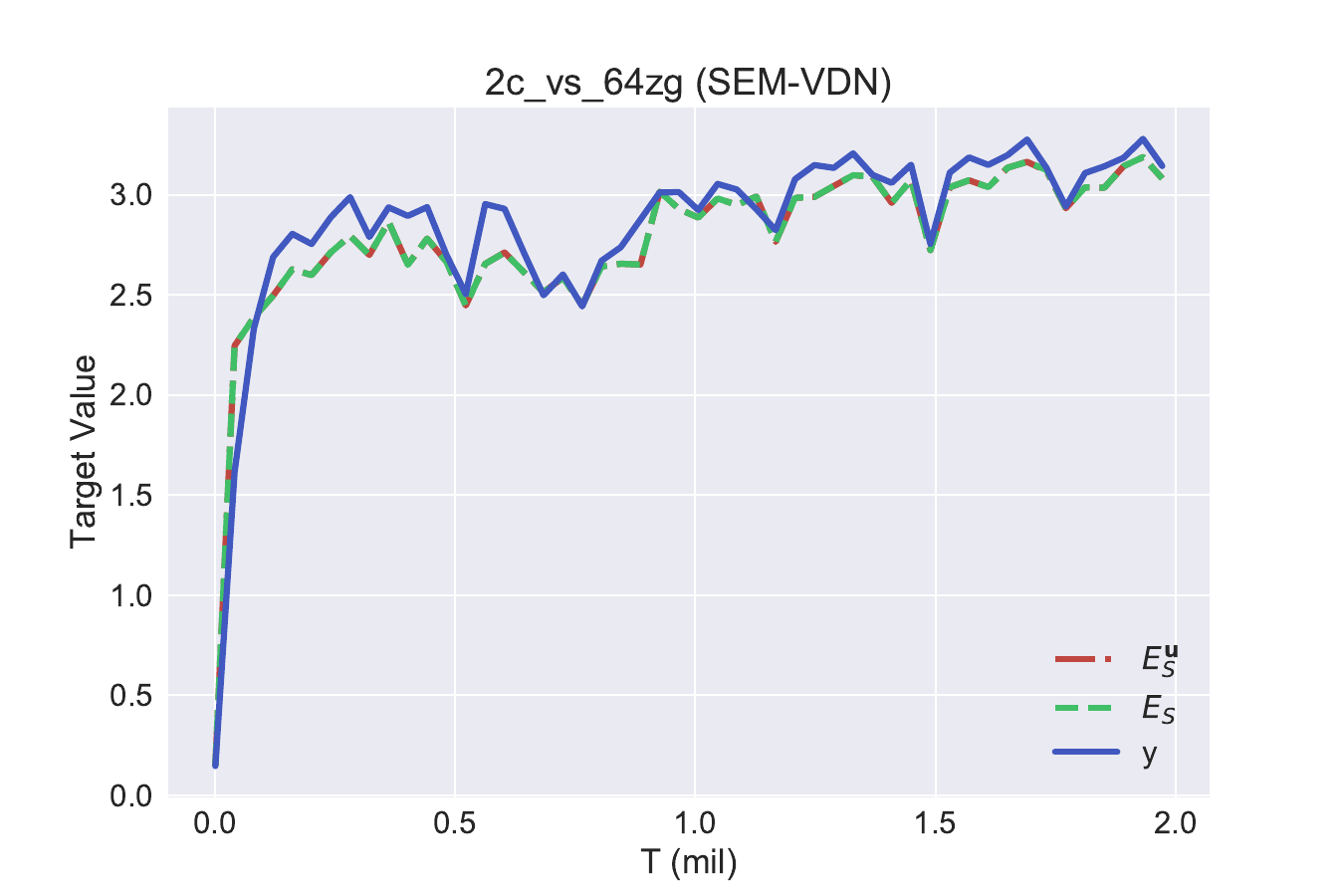}}
\subfigure[SAEM-VDN]{
\includegraphics[width=0.3\linewidth]{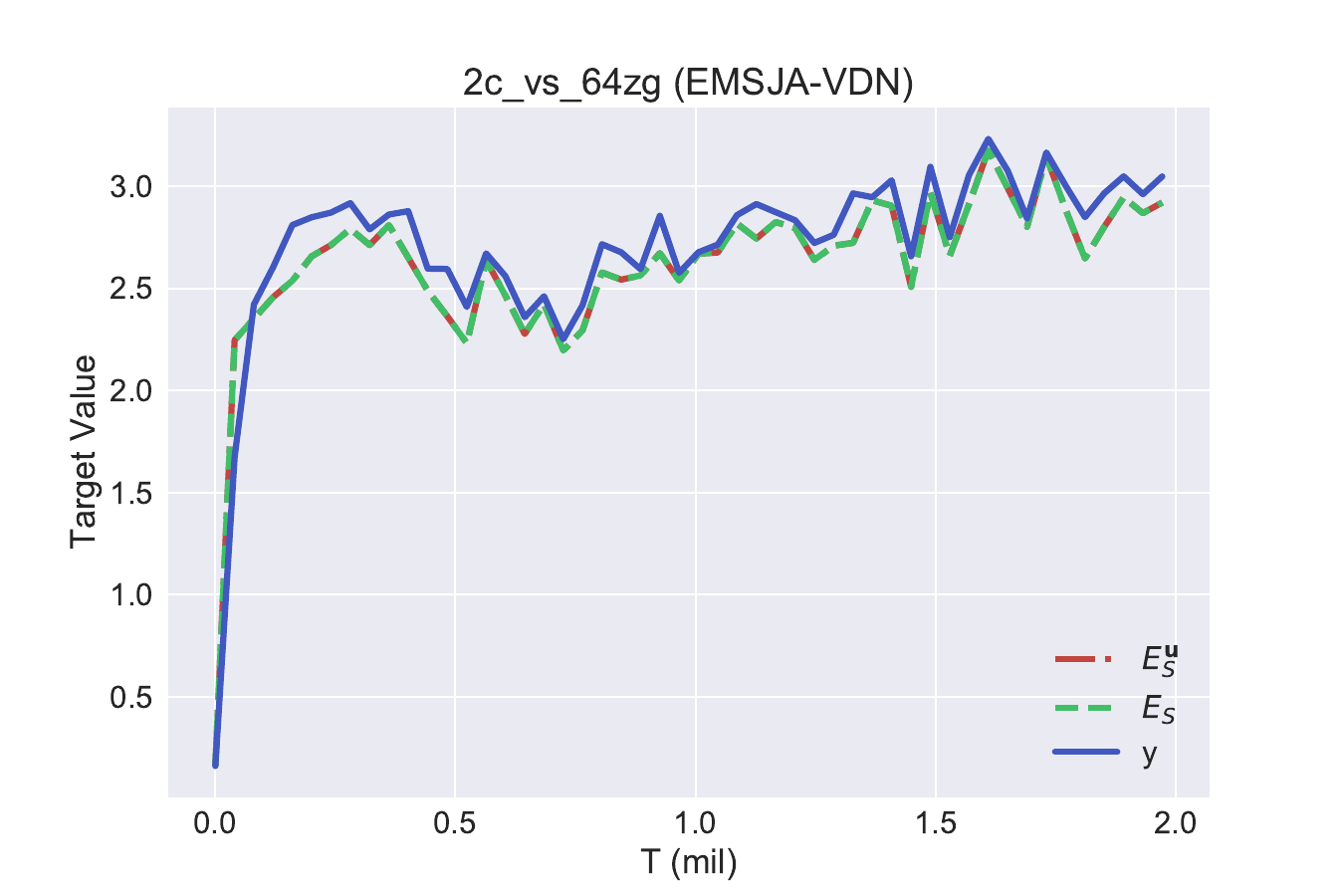}}
\vspace{-0.1cm}
\caption{Comparison among different targets, $E_{s}^{\mathbf{u}}$, $E_{s}$ and $y$, in our methods and baselines. (a) VDN. (b) SEM-VDN. (c) SAEM-VDN. The results are summarized over 5 random runs. For clarity, we represent the median performance without the 25-75\% percentiles.}
\label{compare_target} 
\end{figure*}

To gain the insight of introducing episodic memory into MARL, we try to conduct an in-depth analysis of the training process. We compare different targets in SAEM-VDN, SEM-VDN and VDN on 2c\_vs\_64zg. $y_t^b$ is denoted as $y$, which is a target inferred by the target network. $E_t^b$ is denoted as $E_{s}$, which is an episodic memory target replayed from $Q^{\text{S}}$. $E_{s}^{\mathbf{u}}$ denotes $E_{s_t}^{b,\mathbf{u}_t^b}$, which is replayed from $Q^{\text{SA}}$. The results are shown in~Figure~\ref{compare_target}. Both $E_s$ and $E_s^{\mathbf{u}}$  are stable targets, because they are retrieved from the tables which record the highest return from historical episodes. In VDN, we can see that $y$ is higher than the episodic memory targets $\{E_s, E_s^{\mathbf{u}}\}$ with a large margin. The gap between $y$ and $\{E_s, E_s^{\mathbf{u}}\}$ in SEM-VDN and SAEM-VDN is smaller than that in VDN. It illustrates that the vanilla target $y$, approximated by using target network, is overestimated and the episodic memory target can provide centralized training with a stable target. Hence, our methods can improve the performance and sample efficiency. Furthermore, we can find that the curves of $E_s^{\mathbf{u}}$ and $E_s$ are almost overlapped in Figure~\ref{compare_target}, which illustrates that using $E_s$ to replace $E_s^{\mathbf{u}}$ is reasonable.  

\subsection{Comparison between SEM and SAEM}
Compared with SAEM, it has been theoretically proved in Section~\ref{sec:complexity} that SEM has lower space complexity and time complexity. We choose 2c\_vs\_64zg and 27m\_vs\_30m as the test environment to compare the space complexity and time complexity of SEM and SAEM. 

The space complexity of SEM and SAEM is affected by the number of lookup tables. Here, the size of all lookup tables is set to 1 million, and the dimension $D$ is set to $4$. For SEM, the storage cost is fixed because it has only one lookup table, which needs $0.029$GB storage space to store. For SAEM, the number of lookup tables is equal to the number of joint actions. In 2c\_vs\_64zg, it contains two allied agents, and each agent has 70 available actions. SAEM needs $142$GB storage space to store $70^{2}$ lookup tables. In 27m\_vs\_30m, there are 27 agents, and each agent has 36 actions. SAEM needs $36^{27}\approx 10^{42}$ lookup tables, which needs $3\times10^{40}$GB storage space to store. We can see that the high storage cost makes the implementation of SAEM difficult and even impossible. But SEM has a much lower storage cost than SAEM.

\begin{figure*}[tb]
\centering
\vspace{-0.1cm}
\subfigure[]{
\label{compare:result} 
\includegraphics[width=0.4\linewidth,height=0.19\linewidth]{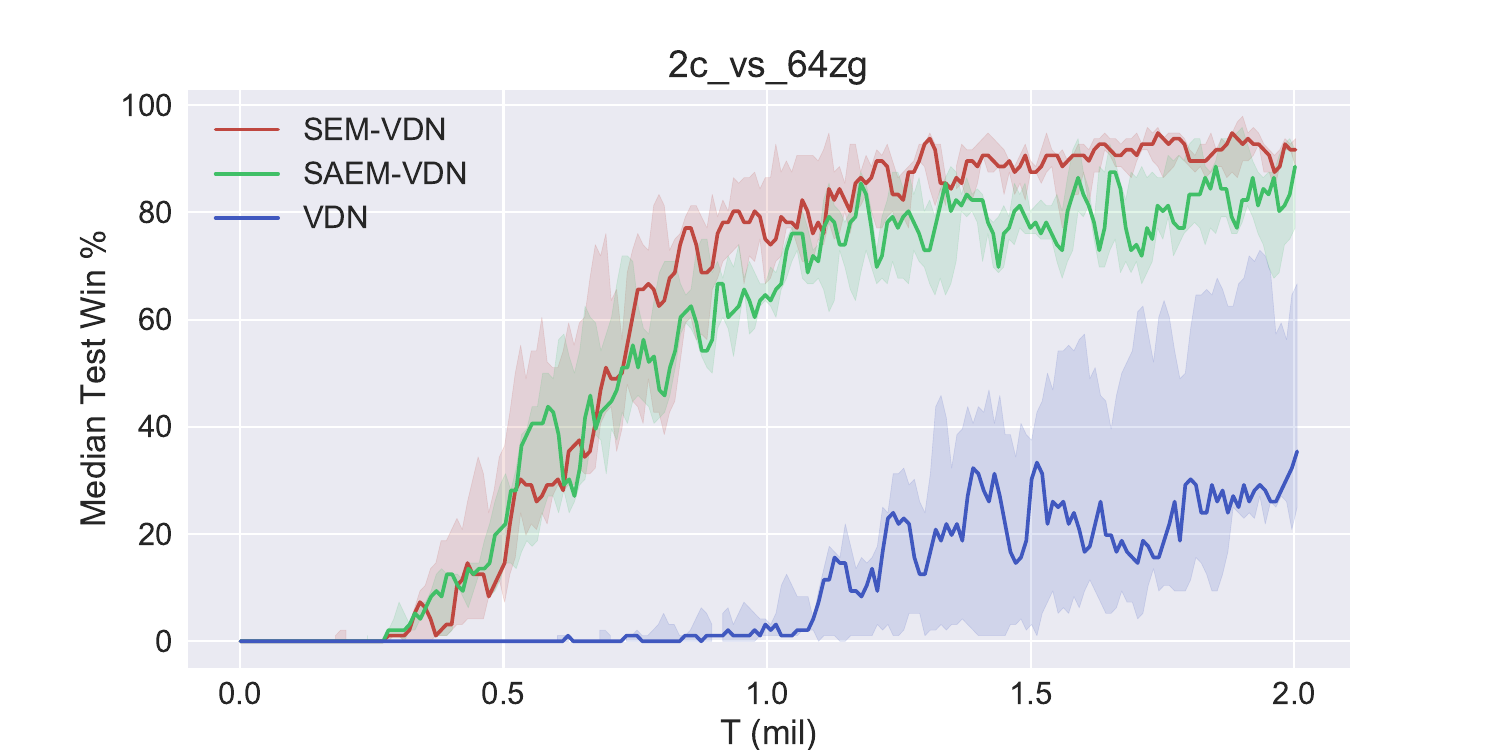}}
\subfigure[]{
\label{compare:time} 
\includegraphics[width=0.4\linewidth,height=0.19\linewidth]{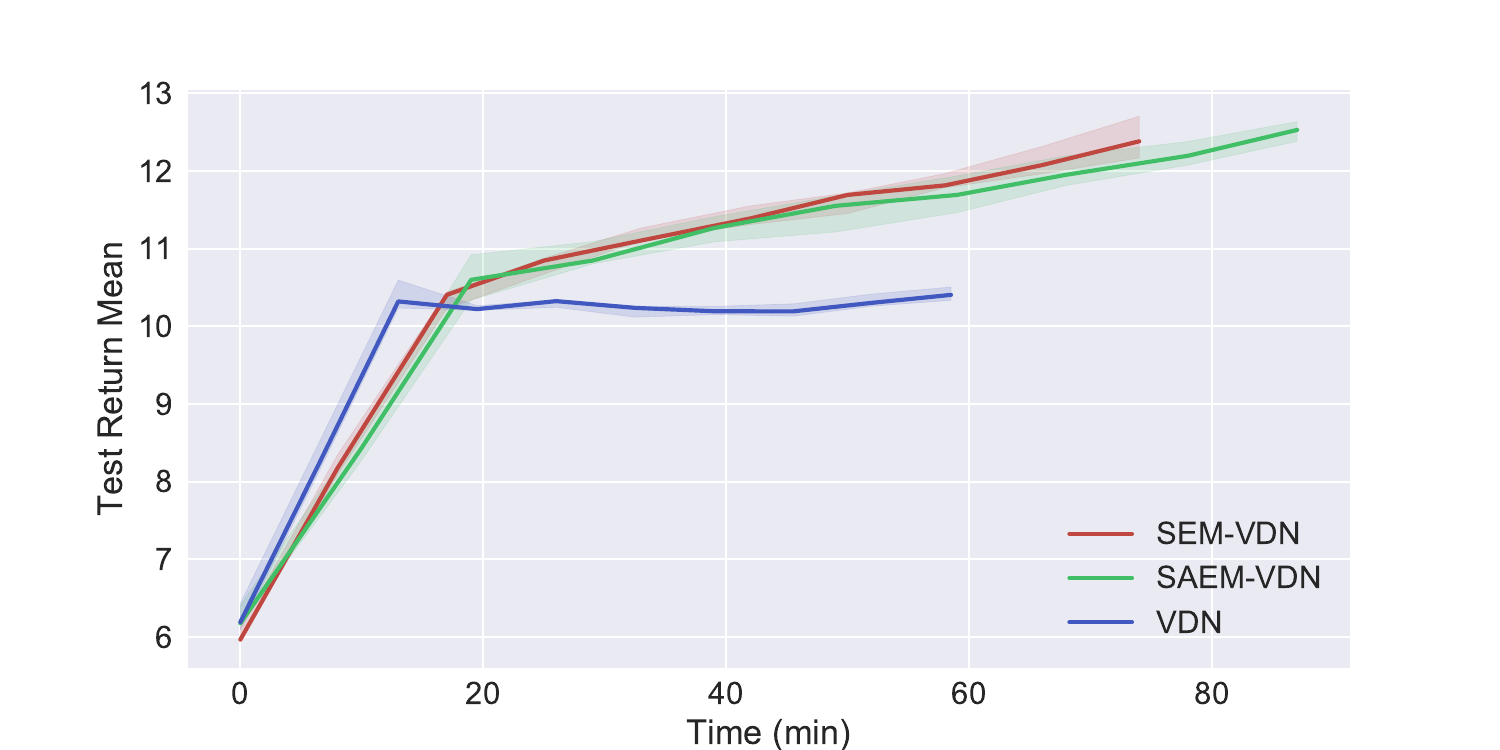}}
\caption{(a) Results of SAEM-VDN, SEM-VDN and VDN on 2c\_vs\_64zg.(b) Test return mean with respect to wall-clock time of SAEM-VDN, SEM-VDN and VDN during the first $100$K time-steps.}
\vspace{-0.8cm}
\end{figure*}
For time cost, we choose 2c\_vs\_64zg to experimentally illustrate that SEM is more time-saving than SAEM. Here, the size of the set $M$ is set to 5K. Other hyper-parameters are described in Appendix. In Figure~\ref{compare:result}, we compare the results of SAEM-VDN, SEM-VDN and VDN. In Figure~\ref{compare:time}, we compare the test return means with respect to the wall-clock time of SAEM-VDN, SEM-VDN and VDN during the first 100K time-steps using Nvidia Geforce GTX 2080 Ti graphics cards. We can see that SEM can use relatively less time than SAEM to achieve a similar return and SEM can save 14\% of time cost compared with SAEM.

\subsection{Learned Policies}
In this section, we investigate the learned behaviors of the different policies in our methods to understand the differences between the strategies better. Here, we choose 27m\_vs\_30m, 2c\_vs\_64zg, MMM2 and bane\_vs\_bane to investigate. For other scenarios, both our methods and baselines can learn a good policy, although the baselines have worse sample efficiency than our methods.

\begin{figure*}[tb]
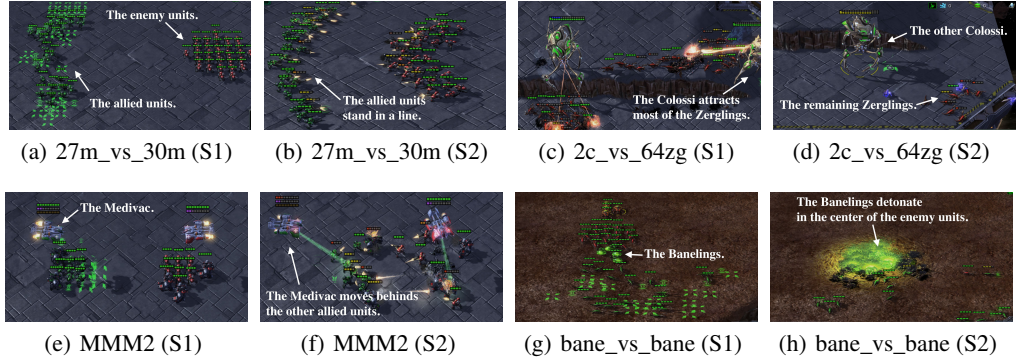

\centering
\subfigure[27m\_vs\_30m~(S1)]{
\label{27m30m:1} 
\includegraphics[width=0.23\linewidth]{27m30m-1.pdf}}
\subfigure[27m\_vs\_30m~(S2)]{
\label{27m30m:2} 
\includegraphics[width=0.23\linewidth]{27m30m-2.pdf}}
\subfigure[2c\_vs\_64zg~(S1)]{
\label{2c64zg:1} 
\includegraphics[width=0.23\linewidth]{2c64zg-1.pdf}}
\subfigure[2c\_vs\_64zg~(S2)]{
\label{2c64zg:2} 
\includegraphics[width=0.23\linewidth]{2c64zg-3.pdf}}
\subfigure[MMM2~(S1)]{
\label{mmm2:1} 
\includegraphics[width=0.23\linewidth]{MMM2-1.pdf}}
\subfigure[MMM2~(S2)]{
\label{mmm2:2} 
\includegraphics[width=0.23\linewidth]{MMM2-3.pdf}}
\subfigure[bane\_vs\_bane~(S1)]{
\label{bane:1} 
\includegraphics[width=0.23\linewidth]{Snapshot-bane_vs_bane-1.pdf}}
\subfigure[bane\_vs\_bane~(S2)]{
\label{bane:2} 
\includegraphics[width=0.23\linewidth]{Snapshot-bane_vs_bane-4.pdf}}
\caption{Illustration of the learned policy in several scenarios.}
\end{figure*}

In the 27m\_vs\_30m scenario, the allied army contains 27 Marines and the enemy army contains 30 Marines. For QMIX, the allied units can learn to stand in a line. For VDN, it bounds to fail the battle. In our methods, the allied army can stand in a line more evenly and more rapidly before the battle than baselines. Hence, these allied units can join the battle faster than those in baselines, as shown in Figure~\ref{27m30m:1} and Figure~\ref{27m30m:2}.

The 2c\_vs\_64zg scenario contains two allied units~(Colossi) but 64 enemy units~(Zerglings), which leads to a much larger action space than the other scenarios. In this scenario, the enemy units are divided into two groups and the allied units are surrounded by enemy units from two opposite directions. For VDN, it fails most of the battles. For QMIX, although it can learn a good policy, it has worse sample efficiency than our methods. For our methods, a Colossi attracts most of the Zerglings to move far away from the other Colossi. The Colossi, with most of the Zerglings, kills some enemy units and then it is killed. The other Colossi kills the enemy units around it and then it searches for the remaining enemy units and kills them.

For MMM2, it contains 1 Medivac, 2 Marauders and 7 Marines. The Medivac can heal damage for Marauders and Marines. Therefore, the key to winning the battle is the Medivac. In our method, the Medivac can move behind the allied units and avoid sacrifice, which can heal other allied units continuously. In baselines, the Medivac can also move behind the allied units, but it is too late in the battle to sacrifice finally and cannot heal other agents continuously.

On the bane\_vs\_bane scenario, it contains a large number of allied and enemy units. The allied army and the enemy army contain 20 Zerglings and 4 Banelings, respectively. Both VDN and QMIX struggle and exhibit large variance, as shown in~Figure~\ref{fig:banebane}. Our methods can learn faster and finally converge. The essentially learned policy of our method is that the four allied Banelings walk into the enemy army's center (Figure~\ref{bane:1}) and then detonate where it is standing, damaging almost all of the enemy units (Figure~\ref{bane:2}). This learned policy is concise and practical, which alleviates the instability to a large extent. 

\subsection{Sensitivity to Hyper-Parameters}
\begin{figure*}[tb]
\centering
\vspace{-0.4cm}
\subfigure[$\lambda$~(SEM-VDN)]{
\label{vdnbanelambda} 
\includegraphics[width=0.4\linewidth]{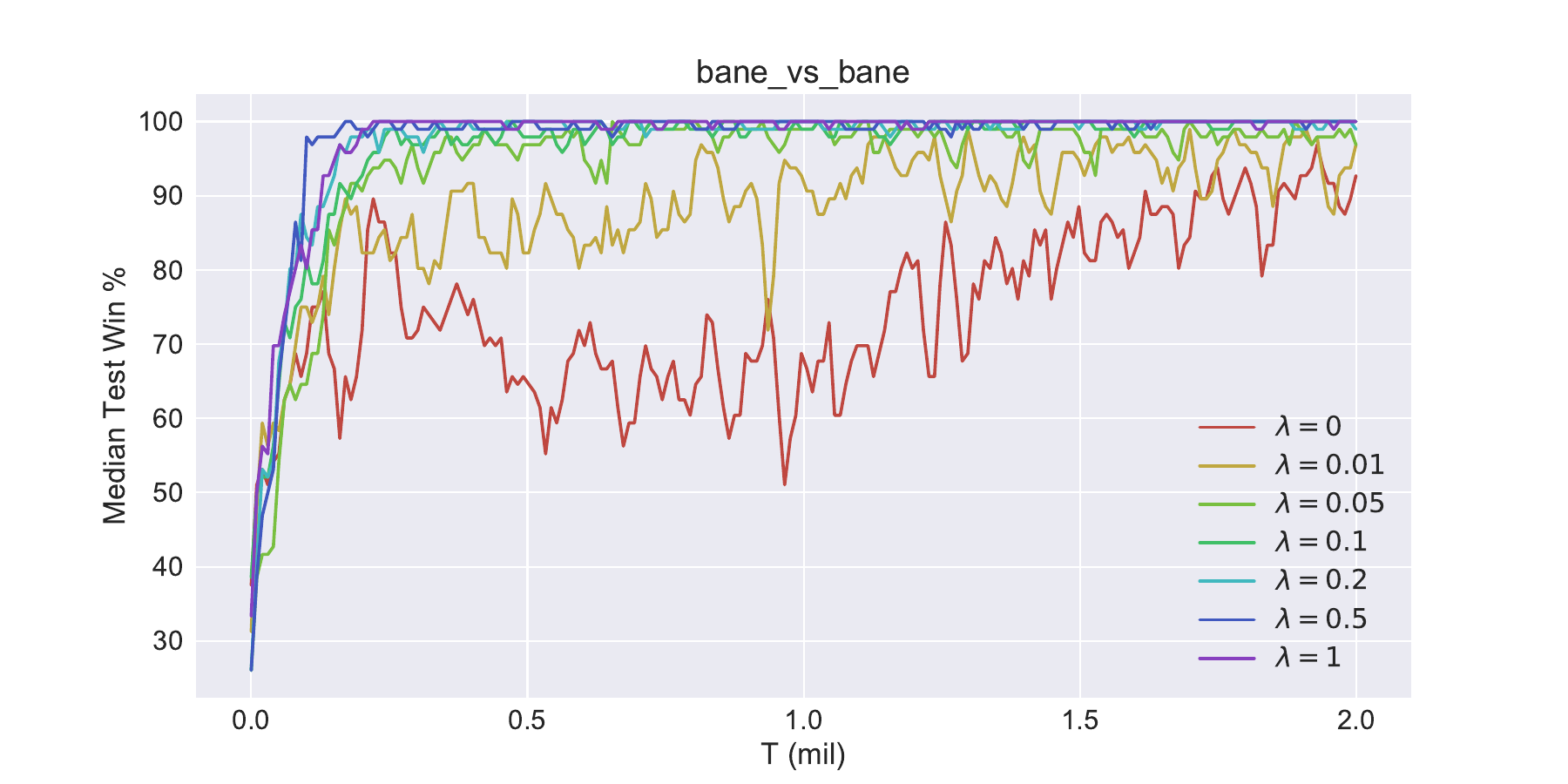}}
\subfigure[$|Q^{\text{S}}|$~(SEM-VDN)]{
\label{vdnbanelambda} 
\includegraphics[width=0.4\linewidth]{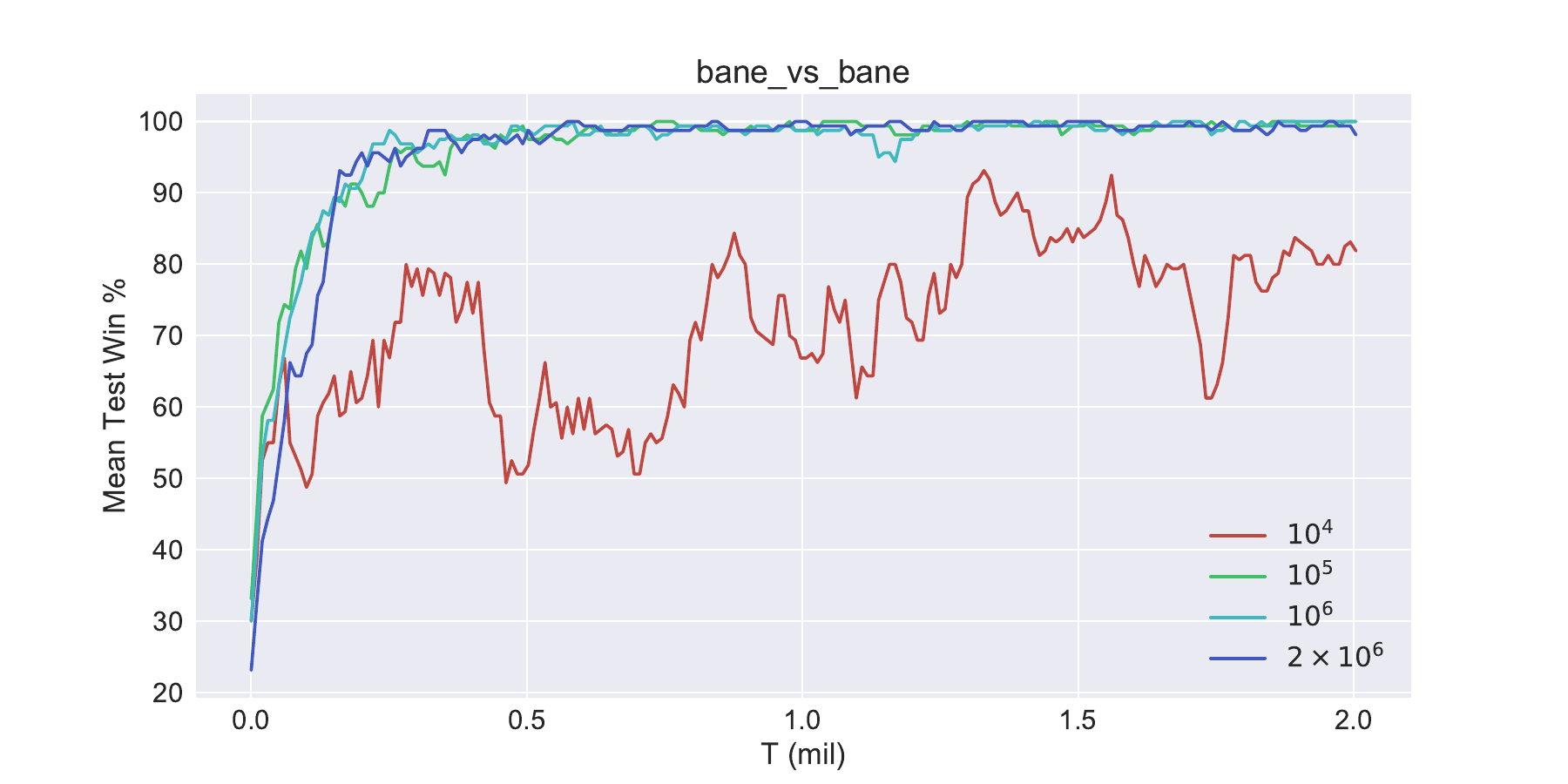}}
\vspace{-0.2cm}
\caption{Median performance of \mbox{SEM}-VDN with different value of the coefficient $\lambda$ and $|Q^{\text{S}}|$. The results are summarized over 5 random runs. For clarity, we represent the median performance without the 25-75\% percentiles.}
\label{compare_lambda}
\vspace{-0.4cm}
\end{figure*}

In order to better understand our method, we investigate the effect of the balance coefficient $\lambda$ and the size of the lookup table $|Q^{\text{S}}|$ on bane\_vs\_bane, shown in Figure~\ref{compare_lambda}. The coefficient $\lambda$ is chosen from $\{0,0.01,0.05,0.1,0.2,0.5,1.0\}$ and $|Q^{\text{S}}|$ is chosen from $\{10^4,10^5,10^6,2\times10^6\}$. When $\lambda$ is set to 0, our method degenerates to the baseline. In most cases, our method performs better than the baseline. When $\lambda=1.0$, our method only uses episodic memory to supervise the training procedure. When $\lambda \in \{0.05,0.1,0.2,0.5,1\}$, our method performs better and learns a good policy faster than the baseline. For $|Q^{\text{S}}|$, we can see that when $|Q^{\text{S}}|$ is small, it would lose some information and then deteriorate the performance. When $|Q^{\text{S}}| \in \{10^5, 10^6, 2\times10^6\}$, our method performs well. We also investigate the sensitivity to the other hyper-parameters in the appendix, including the update frequency of $Q^{\text{S}}$ and the dimension of random projection for state representation.

\section{Conclusion}
In this paper, we have proposed a novel and effective method, SEM, to improve sample efficiency in MARL. To our best knowledge, this is the first work that introduces episodic memory into the multi-agent setting. Compared with the existing EM mechanisms, SEM has lower space complexity and time complexity. Experimental results on SMAC have verified the effectiveness and efficiency of SEM.

\bibliographystyle{abbrv}
\bibliography{neurips_2021_appendix}
\newpage
\appendix

\section{StarCraft Multi-Agent Challenge}
\label{experiment_details}
\subsection{StarCraft Multi-Agent Challenge Setup}
\begin{figure*}[!htb]
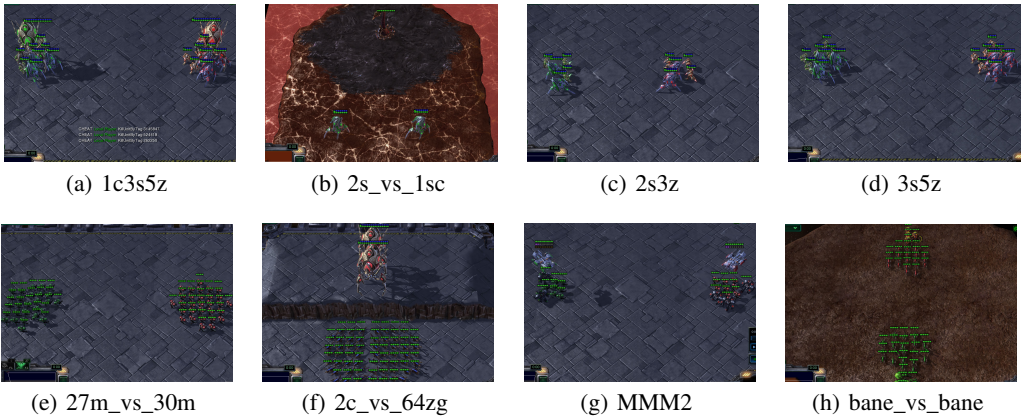

\centering
\subfigure[1c3s5z]{
\label{snap-1c3s5z} 
\includegraphics[width=0.22\linewidth, height=0.15\linewidth]{Snapshot-1c3s5z.pdf}}
\hspace{0.01\linewidth}
\subfigure[2s\_vs\_1sc]{
\label{snap-2s1sc} 
\includegraphics[width=0.22\linewidth, height=0.15\linewidth]{Snapshot-2s_vs_1sc.pdf}}
\hspace{0.01\linewidth}
\subfigure[2s3z]{
\label{snap-2s3z}
\includegraphics[width=0.22\linewidth, height=0.15\linewidth]{Snapshot-2s3z.pdf}}
\hspace{0.01\linewidth}
\subfigure[3s5z]{
\label{snap-3s5z} 
\includegraphics[width=0.22\linewidth, height=0.15\linewidth]{Snapshot-3s5z.pdf}}
\subfigure[27m\_vs\_30m]{
\label{snap-27m} 
\includegraphics[width=0.22\linewidth, height=0.15\linewidth]{Snapshot-27m_vs_30m.pdf}}
\hspace{0.01\linewidth}
\subfigure[2c\_vs\_64zg]{
\label{snap-2c64zg} 
\includegraphics[width=0.22\linewidth, height=0.15\linewidth]{Snapshot-2c_vs_64zg.pdf}}
\hspace{0.01\linewidth}
\subfigure[MMM2]{
\label{snap-MMM2} 
\includegraphics[width=0.22\linewidth, height=0.15\linewidth]{Snapshot-MMM2.pdf}}
\hspace{0.01\linewidth}
\subfigure[bane\_vs\_bane]{
\label{snap-bane} 
\includegraphics[width=0.22\linewidth, height=0.15\linewidth]{Snapshot-bane_vs_bane.pdf}}
\caption{Snapshots of some StarCraft scenarios that we consider.}
\label{SnapshotSMAC}
\end{figure*}

We used the open-source implementations of our baseline algorithms, including VDN~\cite{DBLP:conf/atal/SunehagLGCZJLSL18}, QMIX~\cite{DBLP:conf/icml/RashidSWFFW18}, QPLEX~\cite{DBLP:journals/corr/abs-2008-01062} and WQMIX~\cite{DBLP:conf/nips/RashidFPW20} based on the PyMARL framework~\cite{DBLP:conf/atal/SamvelyanRWFNRH19}. In Section~\ref{Experiments}, we choose eight maps of SMAC, including 1c3s5z, 2s\_vs\_1sc, 2s3z, 3s5z, 27m\_vs\_30m, 2c\_vs\_64zg, MMM2, bane\_vs\_bane, as the test environment. The snapshots of eight maps are shown in Figure~\ref{SnapshotSMAC}, and the configurations of these eight maps are described in the Table~\ref{tab:smac}. The hyper-parameters of SEM are illustrated in Table~\ref{tb_hyper-SEM}. For common hyper-parameters, we adopt the default implementation of PyMARL~\cite{DBLP:conf/atal/SamvelyanRWFNRH19}. The replay buffer $H$ stores episodes and its size is set to $5000$. We sample $B=32$ episodes uniformly from the replay buffer. The neural networks are all trained using RMSprop when the learning rate is set to $5\times 10^{-4}$. After every 200 training episodes, the target networks are updated. During training, each agent $a$ uses $\epsilon$-greedy action selection for exploration. $\epsilon$ is annealed linearly from $1.0$ to $0.05$ over 50K time steps, and then is fixed as a constant. We set $\gamma$ to $0.99$ for all experiments. The enemy agents are controlled by built-in game AI. Depending on the exact scenario, each run of SEM-VDN takes between 11 to 26 hours using an Nvidia Geforce GTX 2080 Ti graphics card.
\begin{table*}[!htb]
\centering
\caption{SMAC challenges}
\begin{tabular}{ccc}
\toprule
Map Name  & Ally Units & Enemy Units \\
\midrule
1c3s5z   & 1 Colossus, 3 Stalkers \& 5 Zealots  & 1 Colossus, 3 Stalkers \& 5 Zealots  \\
2s\_vs\_1sc & 2 Stalkers & 1 Spine Crawler \\
2s3z       & 2 Stalkers \& 3 Zealots  &       2 Stalkers \& 3 Zealots \\
3s5z    & 3 Stalkers \& 5 Zealots  & 3 Stalkers \& 5 Zealots     \\
27m\_vs\_30m & 27 Marines & 30 Marines \\
2c\_vs\_64zg & 2 Colossi & 64 Zerglings\\
MMM2 & 1 Medivac, 2 Marauders \& 7 Marines & 1 Medivac, 2 Marauders \& 8 Marines \\
bane\_vs\_bane & 20 Zerglings \& 4 Banelings & 20 Zerglings \& 4 Banelings \\
\bottomrule
\end{tabular}
\label{tab:smac}
\end{table*}

\begin{table}[htb]
\caption{The hyper-parameters in SEM on SMAC.}
\label{tb_hyper-SEM}
\centering
\begin{tabular}{c|c}
\toprule
  Hyper-parameter  & Value   \\
  \toprule
  Tha balance coefficient $\lambda$ & 0.1\\ 
  The dimension of random projection for state representation        $D$       & 4\\
  The size of lookup table $|Q^{\text{S}}|$ & $10^{6}$\\
  The update frequency of lookup table $|M|$     &      5000         \\
\bottomrule 
\end{tabular}
\end{table} 

\subsection{SEM Combined with Baselines}
We choose several value-decomposed MARL algorithms as baselines, including VDN~\cite{DBLP:conf/atal/SunehagLGCZJLSL18}, QMIX~\cite{DBLP:conf/icml/RashidSWFFW18}, QPLEX~\cite{DBLP:journals/corr/abs-2008-01062} and WQMIX~\cite{DBLP:conf/nips/RashidFPW20}. We combine SEM with these value-decomposed MARL algorithms, respectively denoted as SEM-VDN, SEM-QMIX, SEM-QPLEX, and SEM-WQMIX. 
\subsubsection{SEM-VDN}
The detailed architecture of VDN is illustrated in~\cite{DBLP:conf/atal/SunehagLGCZJLSL18}. For SEM-VDN, the loss is as specified in Eq~(\ref{EM_loss}). The special hyper-parameters of SEM-VDN are illustrated in Table~\ref{tb_hyper-SEM}. The other hyper-parameters of SEM-VDN are the same as that in VDN~\cite{DBLP:conf/atal/SunehagLGCZJLSL18}.
\subsubsection{SEM-QMIX}
The architecture of QMIX has been illustrated in~\cite{DBLP:conf/icml/RashidSWFFW18}. The hyper-parameters of QMIX are the same as that in QMIX~\cite{DBLP:conf/icml/RashidSWFFW18}. In SEM-QMIX, the loss is shown in Eq~(\ref{EM_loss}). The special hyper-parameters of SEM-QMIX are the same as that in Table~\ref{tb_hyper-SEM}.
\subsubsection{SEM-QPLEX}
The overall architecture of QPLEX consists of two key components: a duplex dueling component and an individual action-value function, described as~\cite{DBLP:journals/corr/abs-2008-01062}. In the centralized training, the parameters of the whole network are learned by minimizing TD loss as specified in Eq~(\ref{tderror}). The loss of SEM-QPLEX is as specified in Eq~(\ref{EM_loss}). The special hyper-parameters of SEM-QPLEX are illustrated in Table~\ref{tb_hyper-SEM}. The other hyper-parameters of SEM-QPLEX are the same as that in QPLEX~\cite{DBLP:journals/corr/abs-2008-01062}.

\subsubsection{SEM-WQMIX}
The architecture of WQMIX is described as~\cite{DBLP:conf/nips/RashidFPW20}. The loss of WQMIX is as follows:
\begin{equation}
\begin{aligned}
L(\theta) &= \sum_{b=1}^{B}\sum_{t=1}^{T} w(s_t^b, \mathbf{u}_t^b)\left(Q_{t o t}(\boldsymbol{\tau}_t^b, \mathbf{u}_t^b, s_t^b)-y_t^b\right)^{2} + \sum_{b=1}^{B}\sum_{t=1}^{T}\left(\hat{Q}^{*}(s_t^b, \boldsymbol{\tau}_t^b, \mathbf{u}_t^b)-y_t^b\right)^{2} , \\
\end{aligned}
\label{WQMIX_loss}
\end{equation}
where $y_t^b =r_t^b+\gamma \hat{Q}^{*}\left(s_{t+1}^b, \boldsymbol{\tau}_{t+1}^b, \operatorname{arg}_{\mathbf{u}} Q_{t o t}\left(\boldsymbol{\tau}_{t+1}^b, \mathbf{u}, s_{t+1}^b\right)\right)$. For $w(s_t^b, \mathbf{u}_t^b)$, we use centrally-weighting function~(CW), described as follows:
\begin{equation}
	w(s_t^b, \mathbf{u}_t^b)=\left\{\begin{array}{ll}1 & y_{t}^b>\hat{Q}^{*}\left(s_t^b, \boldsymbol{\tau}_t^b, \hat{\mathbf{u}}^{*}\right) \text { or } \mathbf{u}_t^b=\hat{\mathbf{u}}^{*}, \\ \alpha & \text {otherwise. }\end{array}\right.
\end{equation}

We can combine our method, SEM, with WQMIX, called SEM-WQMIX. The loss of SEM-WQMIX is as follows:
\begin{equation}
\begin{aligned}
L(\theta) &= (1-\lambda)\sum_{b=1}^{B}\sum_{t=1}^{T} w(s_t^b, \mathbf{u}_t^b)\left(Q_{t o t}(\boldsymbol{\tau}_t^b, \mathbf{u}_t^b, s_t^b)-y_t^b\right)^{2}\\
&+ \lambda \sum_{b=1}^{B}\sum_{t=1}^{T} w_e(s_t^b, \mathbf{u}_t^b)\left(Q_{t o t}(\boldsymbol{\tau}_t^b, \mathbf{u}_t^b, s_t^b)-E_t^b\right)^{2} \\
& + (1-\lambda)\sum_{b=1}^{B}\sum_{t=1}^{T}\left(\hat{Q}^{*}(s_t^b, \boldsymbol{\tau}_t^b, \mathbf{u}_t^b)-y_t^b\right)^{2} + \lambda\sum_{b=1}^{B}\sum_{t=1}^{T}\left(\hat{Q}^{*}(s_t^b, \boldsymbol{\tau}_t^b, \mathbf{u}_t^b)-E_t^b\right)^{2}, \\
\end{aligned}
\label{EM_WQMIX_loss}
\end{equation}

where $y_{i}$, $w(s,\boldsymbol{u})$ are the same as the loss of WQMIX and $E_t^b = r_t^b +\gamma Q^{\text{S}}(s_{t+1}^b)$. Similar to $w(s,\boldsymbol{u})$, $w_e(s, \mathbf{u})$ is defined as follows:
\begin{equation}
	w_e(s_t^b, \mathbf{u}_t^b)=\left\{\begin{array}{ll}1 & E_{t}^b>\hat{Q}^{*}\left(s_t^b, \boldsymbol{\tau}_t^b, \hat{\mathbf{u}}^{*}\right) \text { or } \mathbf{u}_t^b=\hat{\mathbf{u}}^{*}, \\ \alpha & \text {otherwise. }\end{array}\right.
\end{equation}
In WQMIX and SEM-WQMIX, $\alpha$ is set to 0.75. The other hyper-parameters are the same as that in~\cite{DBLP:conf/nips/RashidFPW20}. For the hyper-parameters of SEM in SEM-WQMIX, they are illustrated as Table~\ref{tb_hyper-SEM}.

\section{Results of SEM on SMAC}
In Table~\ref{tb_results}, we show the mean and median scores of SEM combined with several value-decomposed methods~(VDN, QMIX, QPLEX, WQMIX) on eight maps. Here, we illustrate the formula of the mean score and the formula of the median score. We denote $\text{Map=\{1c3s5z, 2s\_vs\_1sc, 2s3z, 3s5z, 27m\_vs\_30m, 2c\_vs\_64zg, MMM2, bane\_vs\_bane\}}$. The formula of the mean score and the formula of the median score are shown as follows:
\begin{equation}
	\text{Mean Score}~(t) = \text{Mean}(\{P_{i}^{t}\}_{i\in \text{Map}}), \\
\end{equation}
\begin{equation}
\text{Median Score}~(t) =  \text{Median}(\{P_{i}^{t}\}_{i\in \text{Map}}),
\end{equation}
where $P_i^t$ is the mean test battle won for the map $i$ at $t$ time steps. The mean test won at 0.25M time steps is shown in Table~\ref{tb_0.25M} and the mean battle won at 0.5M time steps is shown in Table~\ref{tb_0.5M}.
\begin{table}[htb]
\caption{The mean test battle won $P^{\text{0.25M}}$ at 0.25M time steps on eight maps of SMAC.}
\label{tb_0.25M}
\centering
\begin{tabular}{l|cccc}
\toprule
  Method                    & 1c3s5z       & 2s\_vs\_1sc     & 2s3z         & 3s5z           \\
                      \midrule
VDN                   & 1\%                         & 66\%                           & 53\%                      & 1\%                       \\
SEM-VDN  & 25\%               & 66\%                  & 71\%            & 24\%             \\\hline
QMIX                  & 34\%                        & 7\%                            & 61\%                      & 30\%                      \\
SEM-QMIX & 70\%               & 28\%                  & 89\%             & 66\%             \\\hline

QPLEX                 & 61\%                        & 99\%                           & 83\%                      & 31\%                                                   \\
SEM-QPLEX         & 60\%                        & 77\%                           & 90\%                      & 77\%                                                   \\\hline

WQMIX               & 17\%                        & 63\%                           & 44\%                      & 23\%                                                    \\
SEM-WQMIX            & 64\%                        & 17\%                           & 89\%                      & 63\%                            \\
\midrule\midrule
Method                & 27m\_vs\_30m  & 2c\_vs\_64zg   & MMM2        & bane\_vs\_bane  \\\hline
VDN                    & 0\%                             & 0\%                             & 0\%                       & 85\%                              \\
SEM-VDN & 0\%                     & 0\%                    & 0\%              & 99\%                     \\\hline
QMIX                 & 0\%                             & 0\%                             & 0\%                       & 32\%                              \\
SEM-QMIX  & 0\%                    & 0\%                    & 0\%              & 99\%                    \\\hline
QPLEX                  & 0\%                             & 0\%                             & 0\%                       & 87\%                              \\
SEM-QPLEX         & 0\%                             & 0\%                             & 0\%                       & 99\%                              \\\hline
WQMIX               & 0\%                             & 0\%                             & 0\%                       & 61\%                              \\
SEM-WQMIX             & 0\% & 0\% &3\% &100\%         \\

\bottomrule         
\end{tabular}
\end{table}

\begin{table}[htb]
\caption{The mean test battle won $P^{\text{0.5M}}$ at 0.5M time steps on eight maps of SMAC.}
\label{tb_0.5M}
\centering
\begin{tabular}{l|cccc}
\toprule
  Method                    & 1c3s5z       & 2s\_vs\_1sc     & 2s3z         & 3s5z           \\
                      \midrule
VDN                   & 14\%          & 94\%         & 81\%          & 14\%          \\
SEM-VDN  & 67\% & 100\% & 89\% & 61\%   \\\hline
QMIX                  & 61\%         & 31\%          & 91\%          & 62\%          \\
SEM-QMIX & 83\% & 69\%  & 91\% & 84\%  \\\hline
QPLEX                 & 87\%          & 99\%           & 98\%          & 80\%         \\
SEM-QPLEX         & 87\%          & 98\%           & 99\%          & 92\%          \\\hline
WQMIX               & 48\%          & 88\%           & 78\%          & 49\%           \\
SEM-WQMIX            & 72\%          & 19\%           & 95\%          & 75\%         \\\hline

\midrule
Method                & 27m\_vs\_30m  & 2c\_vs\_64zg   & MMM2        & bane\_vs\_bane  \\\hline
VDN                   & 0\%          & 0\%           & 0\%          & 60\%           \\
SEM-VDN  & 1\% & 21\% & 0\% & 98\%  \\ \hline
QMIX                  & 5\%          & 0\%         & 1\%          & 24\%           \\
SEM-QMIX  & 6\% & 0\%  & 6\% & 100\% \\\hline
QPLEX                 & 4\%          & 1\%           & 0\%          & 97\%           \\
SEM-QPLEX        & 14\%         & 17\%          & 0\%          & 100\%          \\\hline
WQMIX               & 0\%          & 2\%          & 0\%          & 99\%           \\
SEM-WQMIX             & 5\%          & 35\%          & 15\%         & 100\%\\
\bottomrule         
\end{tabular}
\end{table}
\section{Training Curves of SEM-VDN and SEM-QMIX}
\begin{figure*}[tb]
\centering
\hspace{-0.1\linewidth}
\subfigure[MMM2]{
\label{fig:mmm2} 
\includegraphics[width=0.4\linewidth]{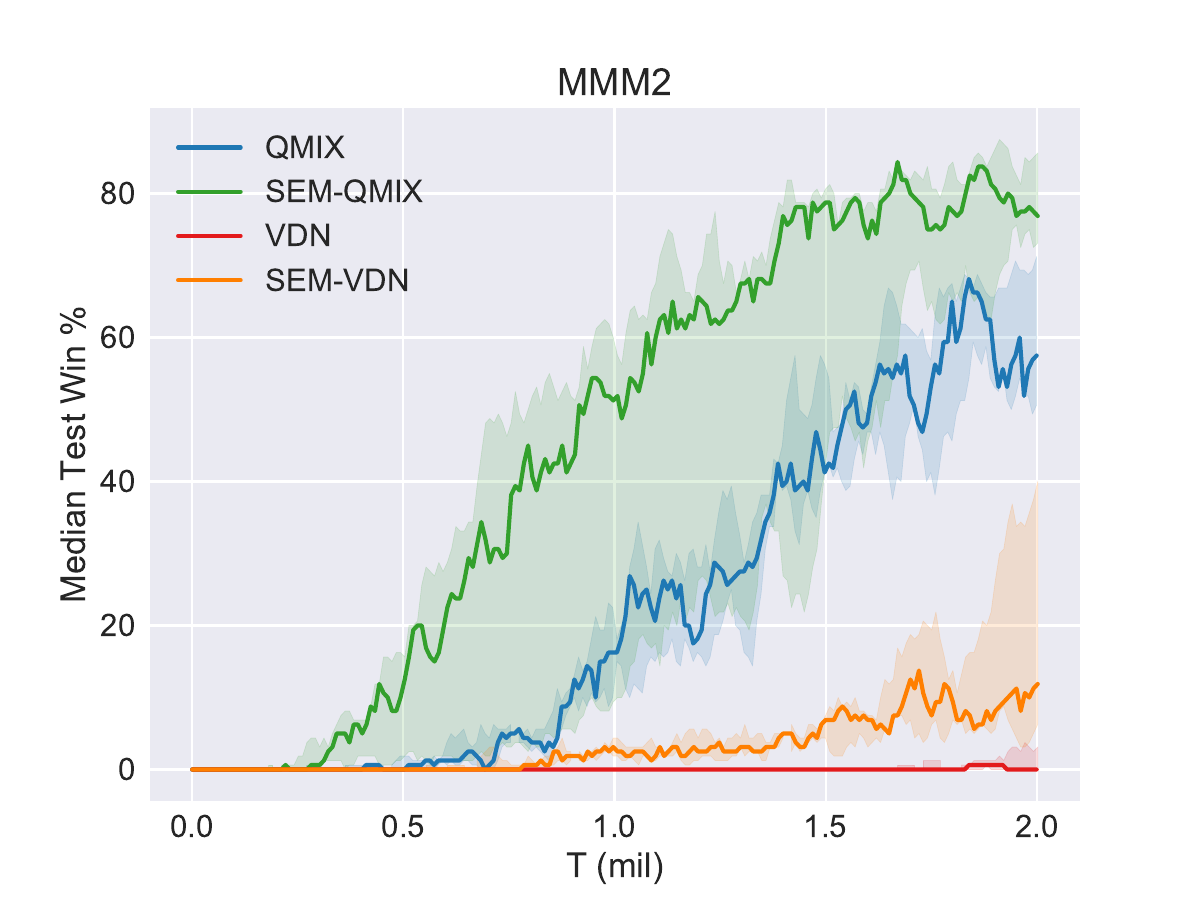}}
\hspace{0.05\linewidth}
\subfigure[2s\_vs\_1sc]{
\label{fig:2s1sc} 
\includegraphics[width=0.4\linewidth]{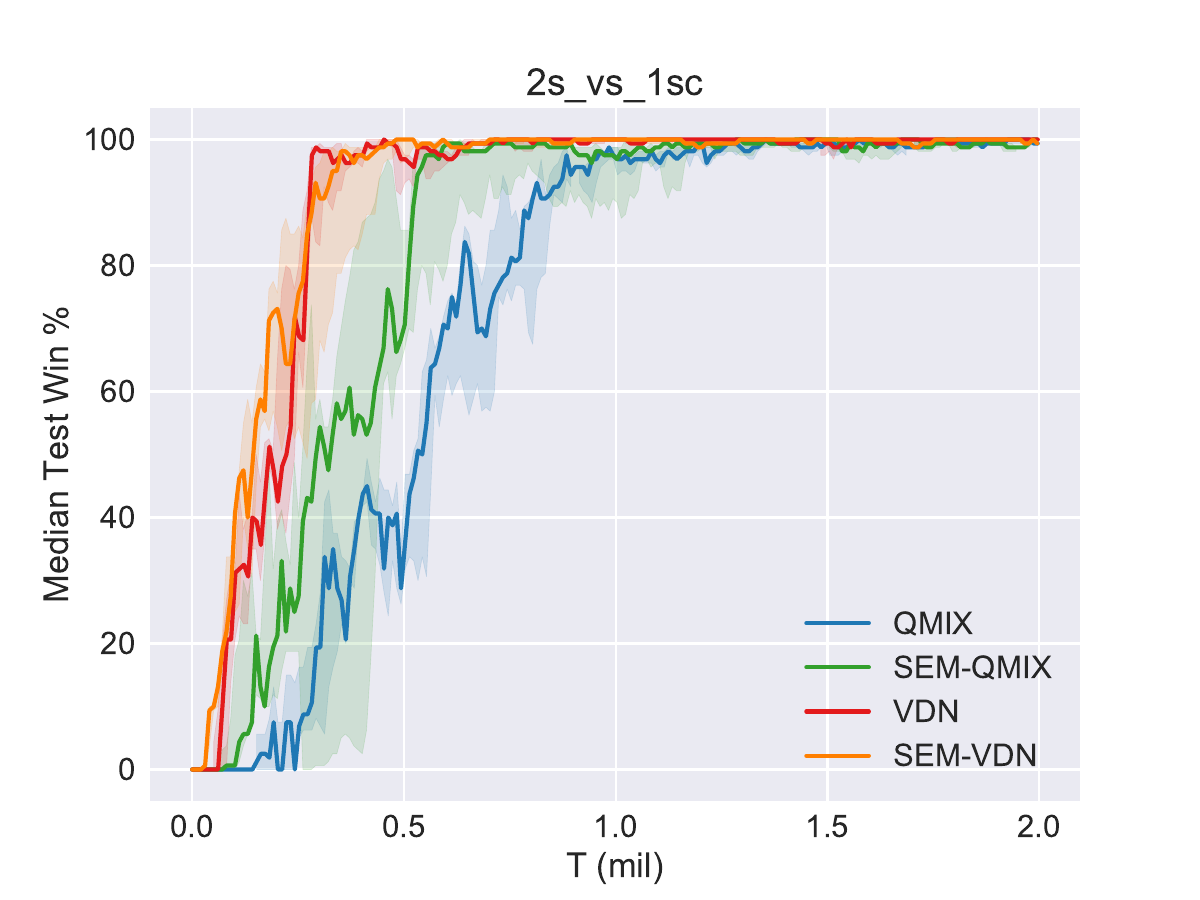}}
\hspace{-0.05\linewidth}
\caption{Results of our methods~(SEM-VDN and SEM-QMIX) and baselines~(VDN and QMIX), including the median performance as well as the 25-75\% percentiles.  }
\vspace{-0.2cm}
\label{EMQMIX2} 
\end{figure*}

In the main paper, we show the training curves of SEM-VDN and SEM-QMIX on six maps. The training curves of SEM-VDN and SEM-QMIX on the other two maps, MMM2 and 2s\_vs\_1sc, are shown in Figure~\ref{EMQMIX2}.
\section{Sensitivity to the Hyper-parameters}
\subsection{Sensitivity to the Hyper-parameter $|M|$}
\begin{figure}[!htb]
\centering
\subfigure[SEM-VDN]{
\label{vdnbanebaneemupdate} \includegraphics[width=0.45\linewidth]{compare_vdn_bane_vs_bane.pdf}}
\subfigure[SEM-QMIX]{
\label{qmixbanebaneemupate} \includegraphics[width=0.45\linewidth]{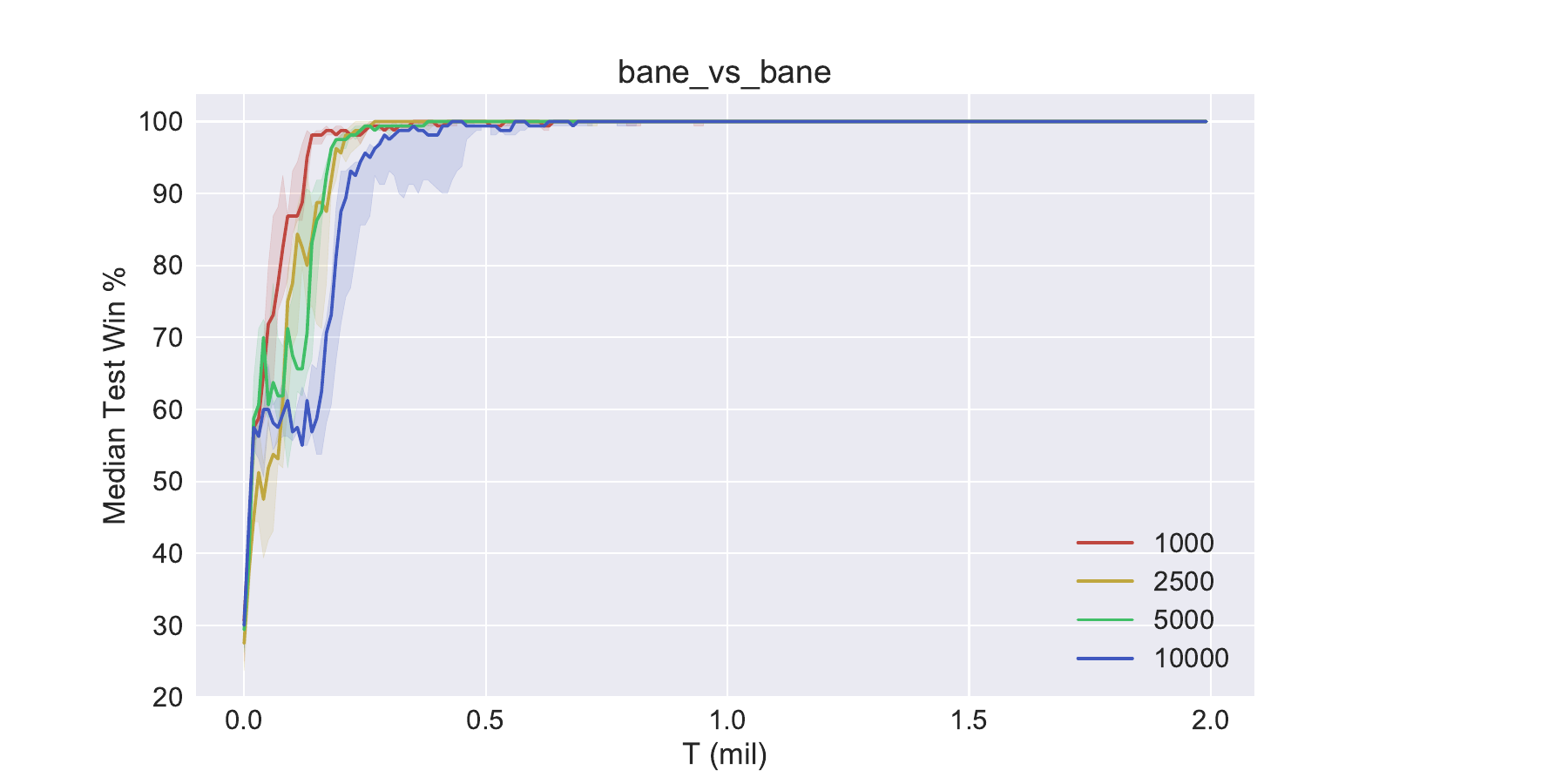}}
\caption{Median performance of SEM-QMIX and SEM-VDN when the episodic memory table is updated after every $|M|$ time steps. $|M|$ is chosen from $\{1000,2500,5000,10000\}$. The results are summarized over 5 random runs.}
\label{compare_emupdate} 
\end{figure}

We choose bane\_vs\_bane to investigate the influence of episodic memory table update frequency. The lookup table $Q^{\text{S}}$ updates when $M$ is filled and then the set $M$ is made empty. In other words, the lookup table updates after every $|M|$ time steps. $|M|$ is chosen from $\{1000, 2500, 5000, 10000\}$. The results are shown in Figure~\ref{compare_emupdate}. We find that a larger $|M|$ might deteriorate the performance because the entries in the lookup table $Q^{\text{S}}$ have not been updated timely with better value. When $|M|\in \{2500,5000\}$, our methods perform better.
\subsection{Sensitivity to the Hyper-parameter $D$}
\begin{figure}[!htb]
\centering
\subfigure[SEM-VDN]{
\label{vdnbanedim} \includegraphics[width=0.45\linewidth]{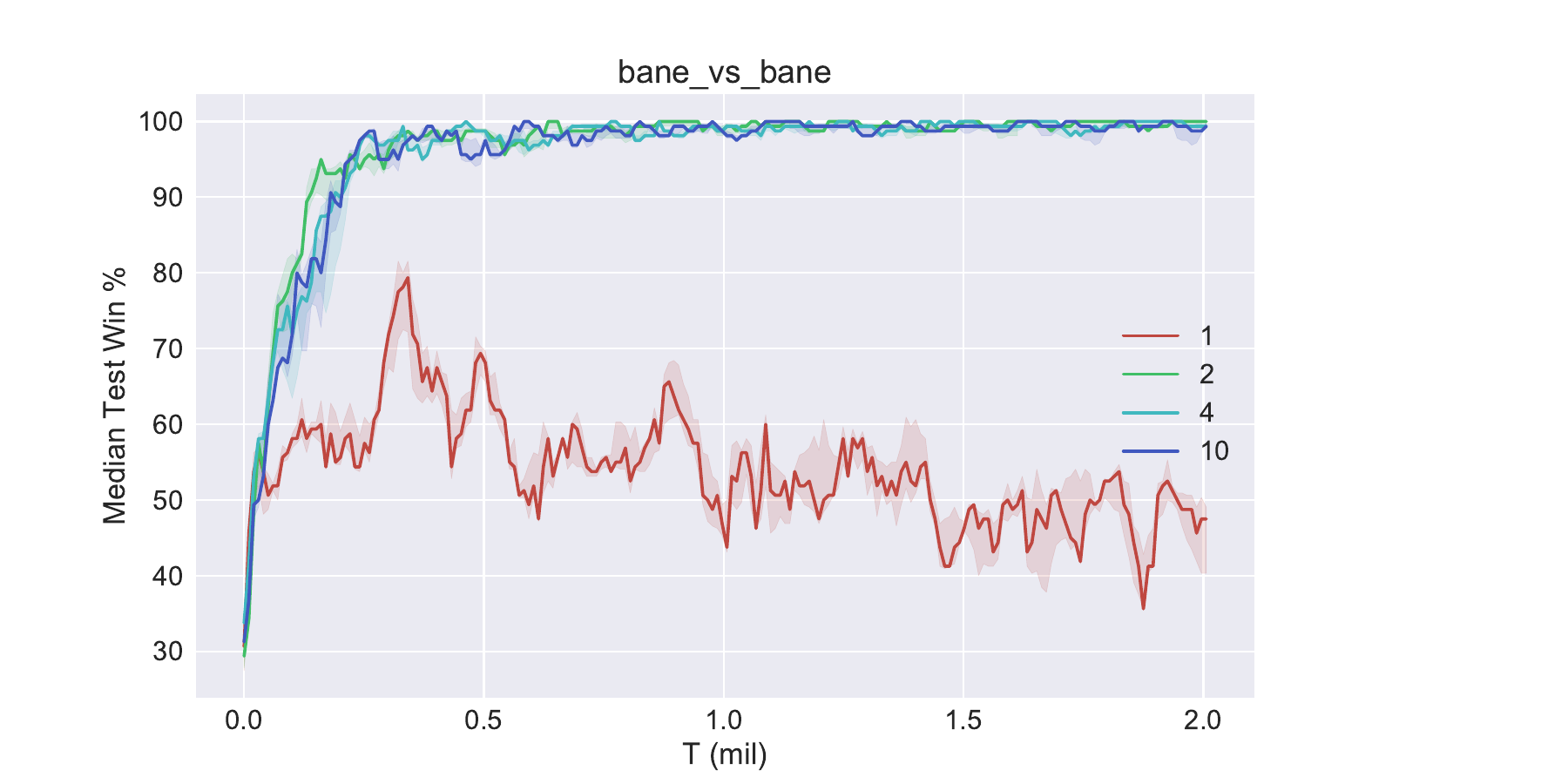}}
\subfigure[SEM-QMIX]{
\label{qmixbanedim} \includegraphics[width=0.45\linewidth]{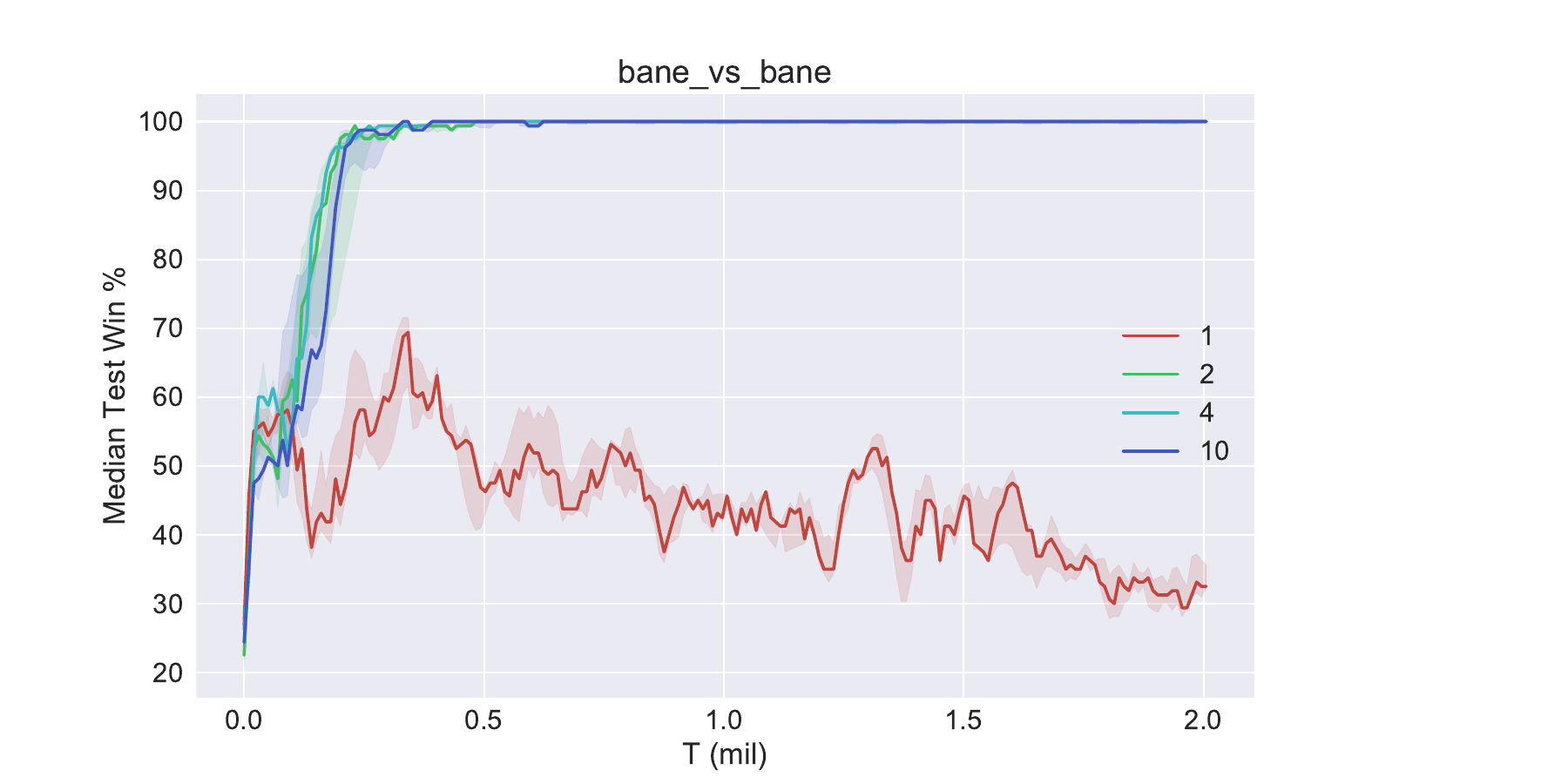}}
\caption{Median performance of SEM-QMIX and SEM-VDN when the dimension of the random projection $D$ is chosen from $\{1,2,4,10\}$. }
\label{compare_dim} 
\end{figure}

We investigate the influence of the dimension of random projection for state representation. $D$ is chosen from $\{1,2,4,10\}$. The results are shown in Figure~\ref{compare_dim}. We can find that if $D$ is too small, performance degrades considerably and if $D$ is too large, the storage overhead is significant. When $D\in\{2,4\}$, it is a reasonable choice.

\end{document}